\documentclass[a4paper,fleqn]{cas-dc}

\usepackage{graphicx}
\usepackage{makecell}
\usepackage{array}
\usepackage{mdwmath}
\usepackage{mdwtab}
\usepackage{eqparbox}
\usepackage{url}
\usepackage{multirow}
\usepackage{amsmath}
\usepackage{caption}
\usepackage{subfigure}
\usepackage{longtable}
\usepackage{algorithmic}
\usepackage{algorithm}
\usepackage{color}
\usepackage{xspace}
\usepackage[numbers]{natbib}

\def\tsc#1{\csdef{#1}{\textsc{\lowercase{#1}}\xspace}}
\tsc{WGM}
\tsc{QE}

\begin{document}
%
\let\WriteBookmarks\relax
\def\floatpagepagefraction{1}
\def\textpagefraction{.001}
\shorttitle{Learning with Limited Annotations: A Survey on Deep Semi-supervised Learning for Medical Image Segmentation}    
\shortauthors{R. Jiao, Y. Zhang et al.} 

\title [mode = title]{Learning with Limited Annotations: A Survey on Deep Semi-supervised Learning for Medical Image Segmentation}  

\author[author1,author9,author4]{Rushi Jiao}
\cormark[1]

\author[author2,author8]{Yichi Zhang}
\cormark[1]

\author[author3]{Le Ding}

\author[author1,author4]{Bingsen Xue}

\author[author3, author7]{Jicong Zhang}

\author[author9,author6 ]{Rong Cai}
\cormark[2]

\author[author1,author4,author5]{Cheng Jin}
\cormark[2]

\address[author1]{ School of Biomedical Engineering, Shanghai Jiao Tong University, Shanghai, 200240, China}

\address[author2]{ School of Data Science, Fudan University, Shanghai, 200433, China}

\address[author9]{ School of Engineering Medicine, Beihang University, Beijing, 100191, China}

\address[author8]{ Artificial Intelligence Innovation and Incubation Institute, Fudan University, Shanghai, 200433, China}

\address[author3]{ School of Biological Science and Medical Engineering, Beihang University, Beijing, 100191, China}

\address[author4]{ Shanghai Artificial Intelligence Laboratory, Shanghai, 200232, China}

\address[author5]{ Beijing Anding Hospital, Capital Medical University, Beijing, 100088, China}

\address[author6]{Key Laboratory for Biomechanics and Mechanobiology of Ministry of Education, Beihang University, Beijing, 100191, China}

\address[author7]{ Hefei Innovation Research Institute, Beihang University, Hefei, 230012, China}

\cortext[1]{Contribute equally to this work.}
\cortext[2]{Corresponding author}

\markboth{  }%
{Shell \MakeLowercase{\textit{et al.}}: Learning with Limited Annotations: A Survey on Deep Semi-Supervised Learning for Medical Image Segmentation}

\begin{abstract}
Medical image segmentation is a fundamental and critical step in many image-guided clinical approaches. Recent success of deep learning-based segmentation methods usually relies on a large amount of labeled data, which is particularly difficult and costly to obtain, especially in the medical imaging domain where only experts can provide reliable and accurate annotations. Semi-supervised learning has emerged as an appealing strategy and been widely applied to medical image segmentation tasks to train deep models with limited annotations. In this paper, we present a comprehensive review of recently proposed semi-supervised learning methods for medical image segmentation and summarize both the technical novelties and empirical results. Furthermore, we analyze and discuss the limitations and several unsolved problems of existing approaches. We hope this review can inspire the research community to explore solutions to this challenge and further advance the field of medical image segmentation.
\end{abstract}


\maketitle

\section{Introduction}
\label{sec:introduction}

\begin{figure}[!t]
    \centering
    \includegraphics[width=\linewidth]{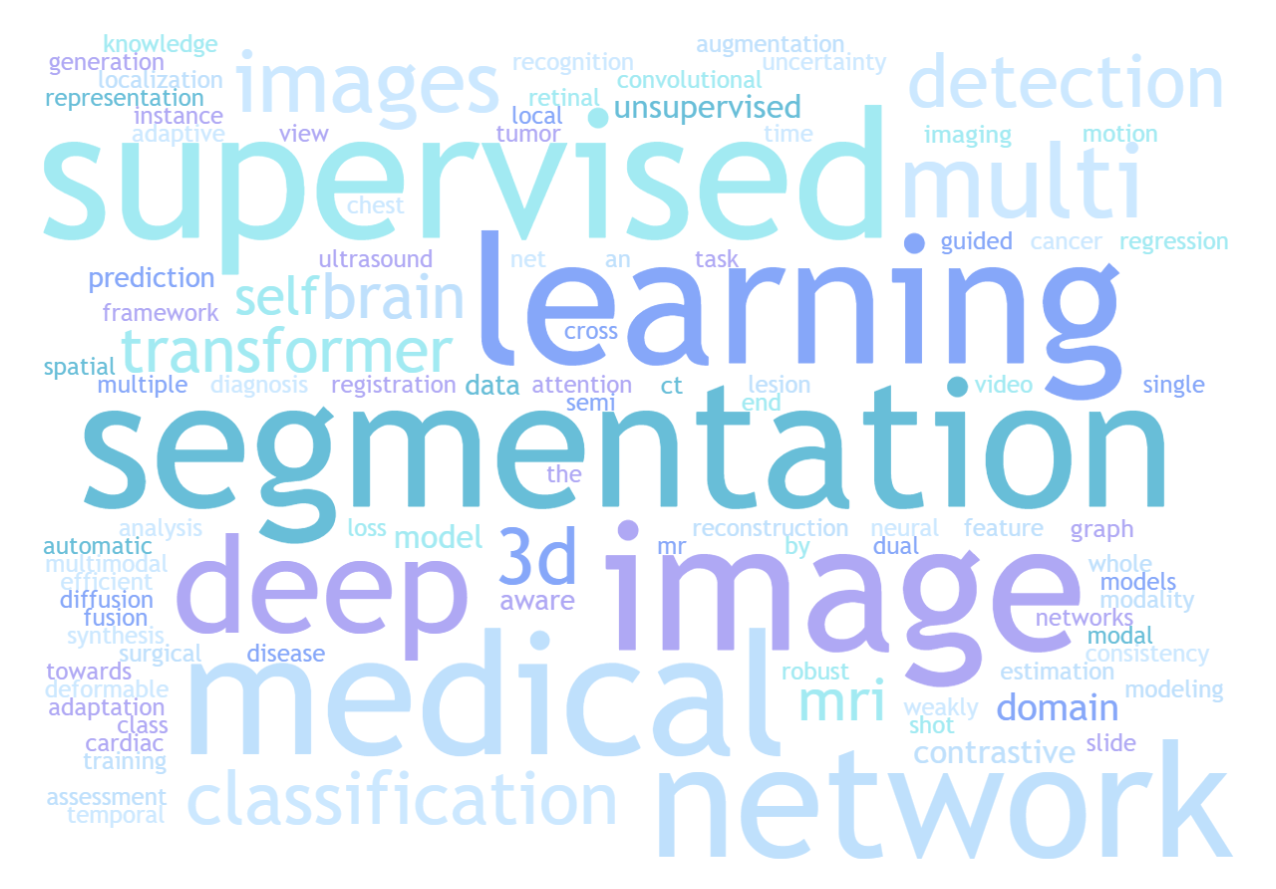}
    \caption{Word cloud of paper titles in MICCAI 2022.}
    \label{wordcloud}
\end{figure}

Medical image segmentation aims to delineate the interested anatomical structures like organs and tumors from the original images by labeling each pixel into a certain class, which is a basic and important step for many clinical approaches like computer-aided diagnosis, treatment planning and radiation therapy \cite{van2011computer,litjens2017survey}.
Accurate segmentation can provide reliable volumetric and shape information so as to assist in further clinical applications like disease diagnosis and quantitative analysis \cite{heller2020state,oreiller2022head,lalande2021deep}.
According to the word cloud of paper titles in the 25rd International Conference and Medical Image Computing and Computer Assisted Intervention \footnote{http://miccai2022.org} (MICCAI 2022) in Figure \ref{wordcloud}, we can observe that "segmentation" is one of the most active research topics and has the highest frequency in medical image analysis community.

Since the introduction of U-Net \cite{ronneberger2015u,cciccek20163d} for medical image segmentation in 2015, many variants of encoder-decoder architecture have been proposed to improve it by re-designing skip connections \cite{zhou2019unet++}, incorporating residual/dense convolution blocks \cite{Alom2019RecurrentRU,li2018h}, attention mechanisms \cite{oktay2018attention,schlemper2019attention,zhang2020sau}, etc. Moreover, nnU-Net (no-new-U-Net) \cite{isensee2020nnunet} can automatically configure the strategies of pre-processing, the network architecture, the training, the inference, and the post-processing to a given dataset for medical image segmentation based on the encoder-decoder structure of U-Net. 
Without manual intervention, nnU-Net surpasses most existing approaches and achieves the state-of-the-art performance in several fully supervised medical image segmentation tasks.
Inspired by recent success of transformer architectures in the field of natural language processing, many transformer-based methods have been proposed and applied for medical image segmentation \cite{chen2021transu,xie2021cotr}.
Although these architectural advancements have shown encouraging results and achieved state-of-the-art performances in many medical image segmentation tasks \cite{Ma2021Cuttingedge3M}, these methods still require relatively large amount of high-quality annotated data for training, more than ever.
However, it is impractical to obtain large-scale carefully-labeled datasets to train segmentation models, particularly for medical imaging where it is hard and expensive to obtain well-annotated data where only experts can provide reliable and accurate annotations \cite{tajbakhsh2020embracing}.
Besides, many commonly used medical images like computed tomography (CT) and magnetic resonance imaging (MRI) scans are 3D volumetric data, which further increase the burden of manual annotation compared with 2D images where experts need to delineate the object from the volume slice by slice \cite{Zhang2020Bridging2A}.

\begin{figure}[!t]
    \centering
    \includegraphics[width=0.8\linewidth]{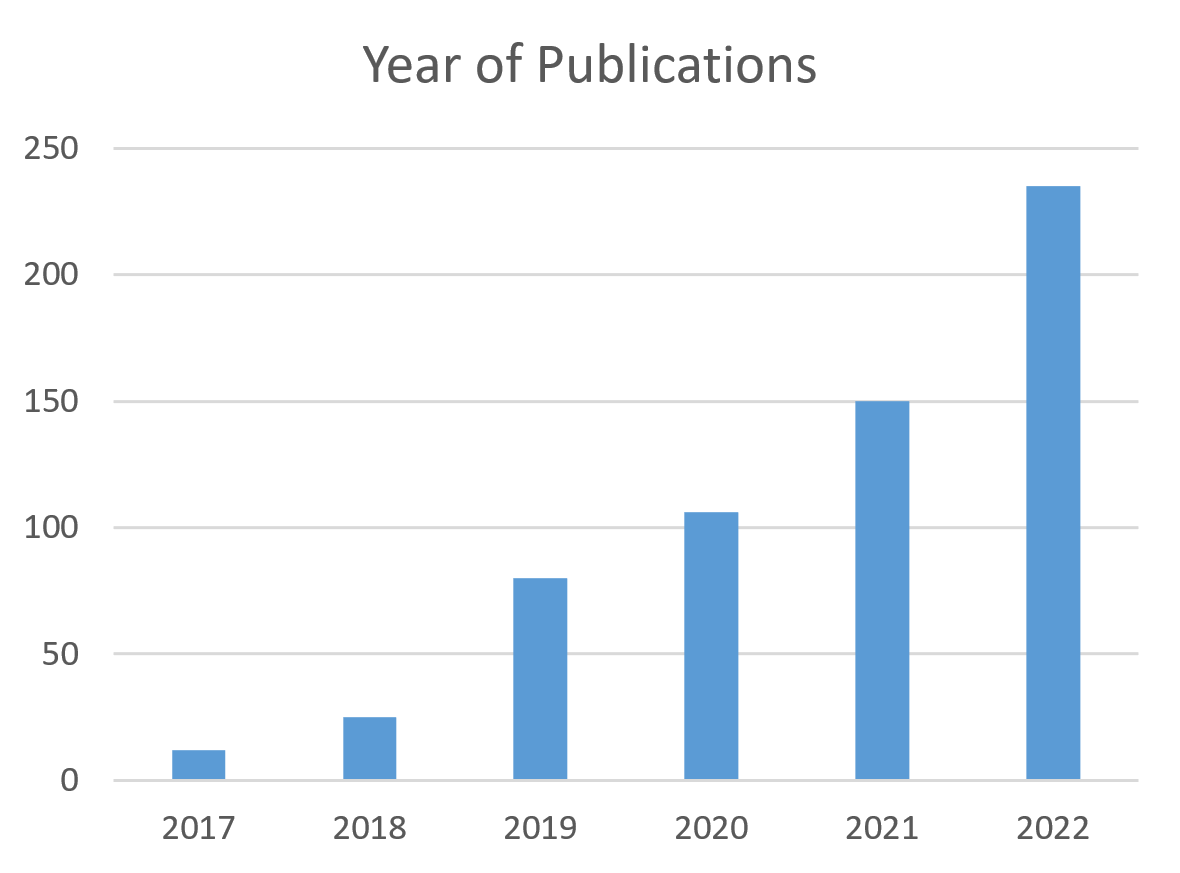}
    \caption{Statistics of papers retrieved from Web of Science on semi-supervised medical image segmentation.}
    \label{statistics}
\end{figure}

To ease the manual labeling burden in response to these challenges, significant efforts have been devoted to annotation-efficient deep learning methods for medical image segmentation tasks by enlarging the training data through label generation\cite{yao2021label}, data augmentation \cite{Zhang2020GeneralizingDL}, leveraging external related labeled datasets \cite{zhang2021exploiting}, and leveraging unlabeled data with semi-supervised learning.
Among these approaches, semi-supervised segmentation is a more practical method by encouraging segmentation models to utilize unlabeled data which is much easier to acquire in conjunction with limited amount of labeled data for training, which has a high impact on real-world clinical applications.
According to the statistics in Figure \ref{statistics}, semi-supervised medical image segmentation has obtained increasing attention from the medical imaging and computer vision community in recent years. 
However, without expert-examined annotations, it is still an open and challenging question on how to efficiently exploit useful information from these unlabeled data.

Main contributions. Compared with related surveys\cite{cheplygina2019not,tajbakhsh2020embracing}, we mainly focus on deep semi-supervised medical image segmentation. And we provide a comprehensive review of recent solutions, summarizing both the technical novelties and empirical results. Furthermore, we analyze and discussed the limitations and several unsolved problems of existing approaches. 
We hope this review could inspire the research community to explore solutions for this challenge and further promote the developments in medical image segmentation field.

\begin{figure}[!t]
    \centering
    \includegraphics[width=9cm]{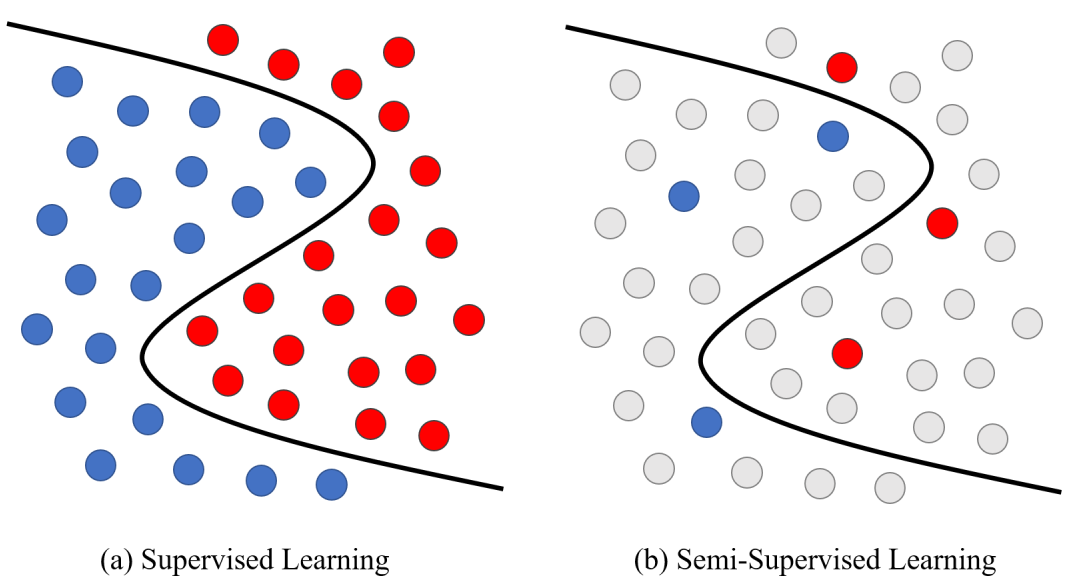}
    \caption{Example comparison of supervised learning and semi-supervised learning.}
    \label{ssl}
\end{figure}

\begin{figure*}[!t]
    \centering
    \includegraphics[width=\textwidth]{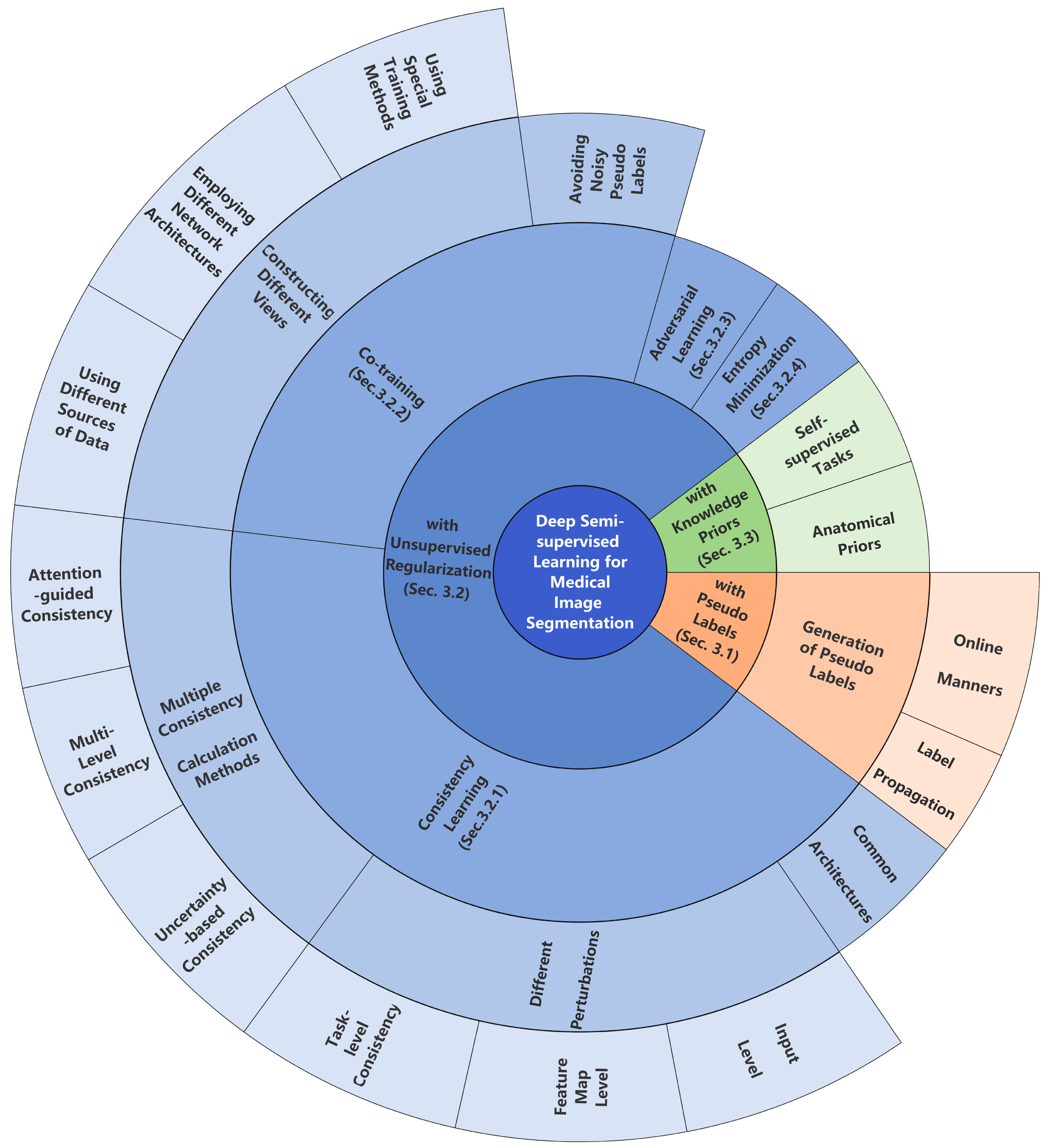}
    \caption{The overview of existing deep semi-supervised learning methods for medical image segmentation.}
    \label{overview2}
\end{figure*}

\section{Preliminaries}

\label{sec:preliminaries}

\subsection{Basic Formulation of Semi-Supervised Learning}

Semi-supervised learning aims to utilize large amount of unlabeled data in conjunction with labeled data to train higher-performing segmentation models. To ease the description in the following sections, we formulate the semi-supervised learning task as follows.

Given a dataset $\mathcal{D}$ for training, we denote the labeled set with $M$ labeled cases as $\mathcal{D}_{L} = \{x_{i}^{l}, y_{i}\}_{i=1}^{M}$, and the unlabeled set with $N$ unlabeled cases as $\mathcal{D}_{U} = \{x_{i}^{u}\}_{i=1}^{N}$, where $x_{i}^{l}$ and $x_{i}^{u}$ denote the input images and $y_{i}$ denotes the corresponding ground truth of labeled data. Generally, $\mathcal{D}_{L}$ is a relative small subset of the entire dataset $\mathcal{D}$, which means $M \ll N$.
For semi-supervised segmentation settings, we aim at building a data-efficient deep learning model with the combination of $\mathcal{D}_{L}$ and $\mathcal{D}_{U}$ and making the performance to be comparable to an optimal model trained over fully labeled dataset.

Based on whether test data are wholly available in the training process, semi-supervised learning can be classified into two settings: the transductive learning and the inductive learning. For transductive learning, it is assumed that the unlabeled samples in the training process are exactly the
data to be predicted (i.e. the test set), and the purpose of the transductive learning is to generalize the model over these unlabeled samples. While for inductive learning, the semi-supervised model will be applied to new unseen data.

\subsection{Assumptions for Semi-Supervised Learning}

For semi-supervised learning, an essential prerequisite is that the data distribution should be under some assumptions. Otherwise, it is impossible to generalize from a finite training set to an infinite invisible set.
The three basic assumptions for semi-supervised learning include \cite{6280908,Yang2021ASO}:

\textbf{The Cluster Assumption.} When two samples $x_{1}$ and $x_{2}$ are similar or belong to the same cluster, their corresponding outputs $y_{1}$ and $y_{2}$ should also be similar or belong to the same category, and vice versa. This assumption implies that the samples in a single class tend to form a cluster. 

\textbf{Low-density Separation.} The decision boundary should be positioned in low-density regions of the feature space rather than high-density regions. This assumption is closely tied to the cluster assumption as it implies that samples belonging to the same class tend to be concentrated in the same cluster. Therefore, a large amount of unlabeled data can be used to adjust the decision boundary.

\textbf{The Manifold Assumption.} If two samples $x_{1}$ and $x_{2}$ reside in a local neighborhood within a low-dimensional manifold, they are likely to possess similar class labels. This assumption reflects the local smoothness of the decision boundary and encourages nearby samples in the feature space to have the same predictions.




\section{Related Work on Semi-Supervised Medical Image Segmentation}
\label{sec:relatedwork}

In this section, we mainly divide these semi-supervised medical image segmentation methods into three strategies as follows:

1) semi-supervised learning with pseudo labels, where unlabeled images
are firstly predicted and pseudo labeled by a segmentation model and then used as new examples for further training.

2) semi-supervised learning with unsupervised regularization, where unlabeled images are used jointly with labeled data to train a segmentation model with unsupervised regularization. This section mainly contains consistency learning, co-training, adversarial learning, entropy minimization.

3) semi-supervised learning with knowledge priors, where unlabeled images is utilized to enable the model with knowledge priors like the shape and position of the targets to improve the representation ability for medical image segmentation.

\begin{table*}
        
	\caption{The summarized review of semi-supervised medical image segmentation methods with pseudo labels.} \label{table_pseudo}
	\centering
	\renewcommand\arraystretch{1.4}
        \scalebox{0.9}{
	\begin{tabular}{l|l|p{1.5cm}|p{4cm}|p{6cm}}
	    \hline
        Reference &
          2D/3D &
          Modality &
          Dataset &
          Label generation methods \\
         \hline
        PLRS, Thompson \textit{et al.} \cite{thompson2022pseudo} &
          3D &
          MRI &
          BraTS 2020 \cite{menze2014multimodal} &
          Superpixel maps calculated by simple linear iterative clustering (SLIC) algorithm \cite{achanta2012slic} to refine pseudo labels \textbf{[online]}.
           \\
        SSA-Net, Wang \textit{et al.}
        \cite{wang2022ssa}&
          2D &
          CT &
          COVID-19-CT-Seg dataset \cite{ma2021toward}, COVID-19 CT Segmentation dataset \footnotemark[1] &
          Add a trust module to re-evaluate the pseudo labels from the model outputs \textbf{[online]}.  \\
        CoraNet, Shi \textit{et al.} \cite{shi2021inconsistency}&
          2D/3D &
          CT, MRI &
          Pancreas CT \cite{clark2013cancer}, MR Endocardium \cite{andreopoulos2008efficient}, ACDC \cite{bernard2018deep} &
          Conservative-radical network to generate more reliable results \textbf{[online]}.  \\
        ECLR, Zhang \textit{et al.}
        \cite{ZHANG2022102458}&
          2D &
          Microscope &
          Gland Segmentation Challenge dataset \cite{sirinukunwattana2017gland}, ColoRectal Adenocarcinoma Gland (CRAG)\cite{Awan20171} &
          Add an error prediction network to divide segmentation errors into intra-class inconsistency or inter-class similarity problems \textbf{[online]}.  \\
        SECT, Li \textit{et al.} \cite{li2021self} &
          2D &
          CT &
          UESTC-COVID-19 Dataset\cite{wang2020noise}, COVID-19-CT-Seg dataset \cite{ma2021toward} &
          Self-ensembling strategy to build the up-to-date predictions via exponential moving average \textbf{[online]}.  \\
        LoL-SSL, Han \textit{et al.}\cite{9757875} &
          2D &
          CT &
          part of LiTS dataset\cite{bilic2019liver} &
          Generate class representations from labeled data based on prototype learning \textbf{[label propagation]}. \\
        NM-SSL, Wang \textit{et al.}
        \cite{2021Neighbor} &
          2D &
          \makecell[l]{X-Ray,\\Dermoscopic} &
          \makecell[l]{ISIC Skin \cite{codella2019skin}, Chexpert \cite{irvin2019chexpert}} &
          Neighbor matching to generate pseudo-labels on a weight basis according to the embedding similarity with neighboring labeled data \textbf{[label propagation]}. \\
        RPG, \textit{Seibold et al.}
        \cite{seibold2021reference} &
          2D &
          X-Ray &
          JSRT dataset\cite{shiraishi2000development} &
          Generate pseudo labels through transferring semantics \textbf{[label propagation]}. \\
		\hline
	\end{tabular}}
	\\
	\footnotetext[1]  01. https://medicalsegmentation.com/covid19/
\end{table*}

\subsection{Semi-Supervised Medical Image Segmentation with Pseudo Labels}

To utilize unlabeled data, a direct and intuitive method is assigning pseudo annotations for unlabeled images, and then using the pseudo labeled images in conjunction with labeled images to update the segmentation model.
Pseudo labeling is commonly implemented in an iterative manner therefore the model can improve the quality of pseudo annotations iteratively. 
Algorithm \ref{Algorithm1} presents the overall workflow of this strategy. 

Firstly, an initial segmentation model is trained using limited labeled data. The initial segmentation model is then applied to unlabeled data to generate pseudo segmentation masks. After that, pseudo-labeled dataset is then merged with labeled dataset to update the initial model. The training procedure alternates between the two steps introduced above, until a predefined iteration number.

\renewcommand{\algorithmicrequire}{ \textbf{Input:}}     
\renewcommand{\algorithmicensure}{ \textbf{Output:}}    
\begin{algorithm}
	\caption{Training procedure of semi-supervised learning with pseudo labels.}
	\label{Algorithm1}
	\begin{algorithmic}[1]
		\REQUIRE{ $\{x^{l}, y^{l}\}$ from labeled dataset $D_{L}$, $\{x^{u}\}$ from unlabeled dataset $D_{U}$, initial segmentation model $\mathcal{M}_{0}$, iteration times $\mathcal{T}$}
		\ENSURE{Trained segmentation model $\mathcal{M}_{\mathcal{T}}$}
		\STATE Training initial segmentation model $\mathcal{M}_{0}$ with $D_{L}$
		\FOR{$i \leftarrow 1$ to $\mathcal{T}$}
		\STATE Generate pseudo labels $\{\hat{y}^{u}\}$ of unlabeled cases $\{x^{u}\}$ with model $\mathcal{M}_{i-1}$
		\STATE Generate new training dataset $D_{PLi}$ with the combination of labeled dataset $\{x^{l}, y^{l}\}$ and pseudo labeled dataset with $\{x^{u}, \hat{y}^{u}\}$
		\STATE $\mathcal{M}_{i}$ $\leftarrow$ Fine-tuning model $\mathcal{M}_{i-1}$ using $D_{PLi}$
		\ENDFOR
		\RETURN {Updated model $\mathcal{M}_{\mathcal{T}}$}
	\end{algorithmic}
\end{algorithm}

Within this strategy for semi-supervised learning, these methods mainly differ in the model initialization, generation of pseudo labels, and how the noise in pseudo labels is handled. The outputs of an under-trained segmentation model with limited labeled data are noisy. If these noisy outputs are used as pseudo labels directly, it may make the subsequent training process unstable and hurt the performance \cite{lee2013pseudo}. For better leverage of the pseudo labels with potential noise, lots of methods have been proposed. In this section, we will explain the generation of pseudo-labels from two aspects: online generation followed by removing noisy predictions and label propagation.

\textbf{Online generation} Pseudo labels are mostly generated through the predictions of a trained model in an online manner followed by some post-processing algorithms for refinement. A common method is to choose unlabeled pixels with maximum predicted probability greater than the setting threshold. However, the predictions can be noisy and unreliable, and may provide incorrect guidance. It is unreasonable to set the same threshold fit for all the samples. In \cite{zeng2023ss}, double-threshold pseudo labeling is introduced, in which predictions from the classification branch and the segmentation branch jointly determine the reliable pseudo labels. Based on the work in \cite{wang2022ssa,lee2013pseudo}, the pseudo labels with higher confidence are usually more effective. Therefore, many confidence- or uncertainty-aware methods are proposed to generate more stable and reliable pseudo labels. For example, Yao \textit{et al.} \cite{yao2022enhancing} propose a confidence-aware cross pseudo supervision network to improve the pseudo label quality of unlabeled images from unknown distributions. Specifically, the input image from source domain is perturbed with the amplitude of the target domain through the Fourier transformation to generate the transformed image. The pixel-wise KL-divergence of the predictions of the original and transformed images is calculated as the variance $V$, which is then used to calculate the pixel-wise confidence. Pseudo labels with high confidence are selected for loss calculation. This process is shown as follows:
\begin{equation}\label{KL-divergence}
    \begin{split}
        V =E[P_{F} \log(\frac {P_{F}}{P_{O}})]
    \end{split}
\end{equation}

\begin{equation}\label{pixel-wise-confidence}
    \begin{split}
        confidence = e^{-V}
    \end{split}
\end{equation}
where, $P_{F}$ and $P_{O}$ represent the predictions of the transformed images and original images. 
Wang \textit{et al.} \cite{wang2022ssa} add a trust module to re-evaluate the pseudo labels from the model outputs and set a threshold to choose high confidence values. Except adding confidence-aware modules, Li \textit{et al.} \cite{li2021self} propose a self-ensembling strategy to build the up-to-date predictions via exponential moving average to avoid noisy and unstable pseudo labels. For post-processing algorithms, morphological methods, machine learning methods \cite{achanta2012slic} and additional networks \cite{shi2021inconsistency,ZHANG2022102458} are usually used to further refine pseudo labels. For example, superpixel maps calculated by simple linear iterative clustering (SLIC) algorithm \cite{achanta2012slic} are introduced to refine pseudo labels in \cite{thompson2022pseudo}. This algorithm is suitable for segmentation of targets with irregular shapes. Shi \textit{et al.} \cite{shi2021inconsistency} propose a conservative-radical network. The object conservative setting tends to predict pixels into background while the object radical setting tends to predict pixels into foreground. The certain region in predictions of unlabeled data is the overlap between conservative and radical settings and employed as pseudo labels. Zhang \textit{et al.} \cite{ZHANG2022102458} rectify the segmentation results of unlabeled data through another error segmentation network followed by the main segmentation network. The segmentation errors are divided into intra-class inconsistency or inter-class similarity problems. This method is applicable for different segmentation models and tasks. Recently, vision foundation models such as Segment Anything Model (SAM) \cite{SAM-Meta}, have shown their amazing capabilities and generalization abilities. \cite{li2023segment} hypothesized that reliable pseudo-labels usually make SAM \cite{SAM-Meta} conduct predictions consistent with the SSL models. So predictions of the SSL models are used as prompts to the SAM \cite{SAM-Meta} to select reliable pseudo-labels. Then the SSL models are retrained with the reliable sets. This method shows a  superior performance compared with existing SSL algorithms.    

\begin{figure*}[!t]
    \centering
    \includegraphics[width=\linewidth]{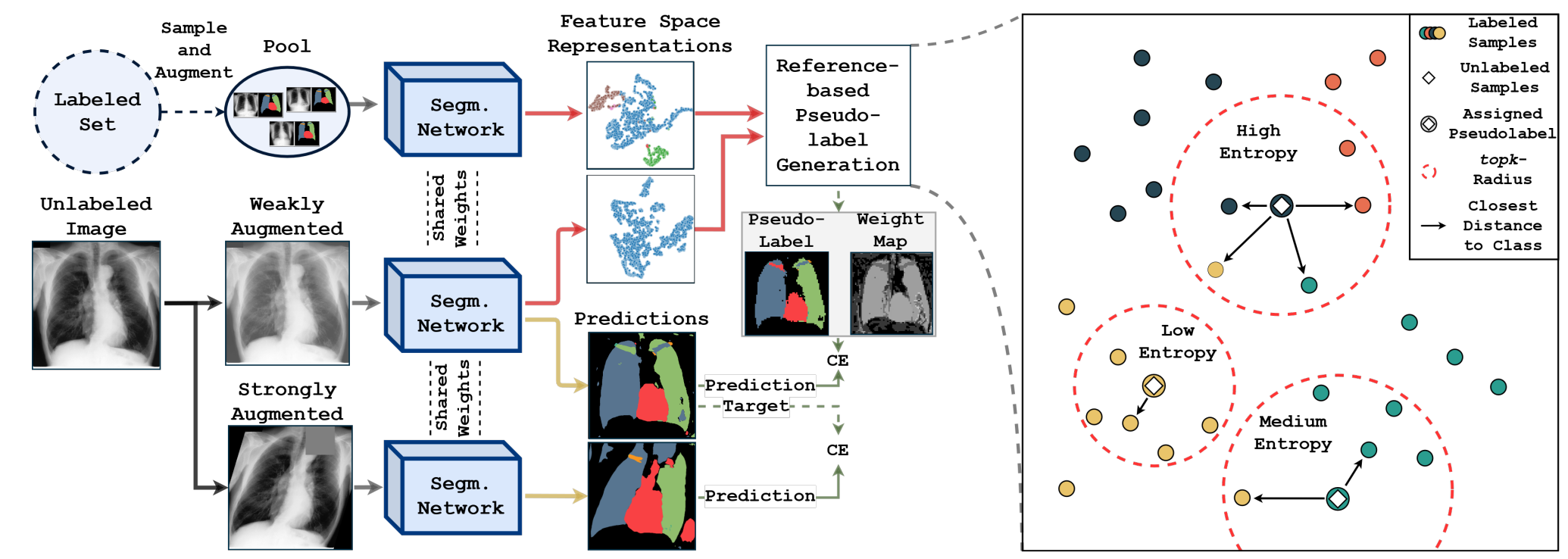}
    \caption{Reference-guided pseudo-label generation \cite{seibold2021reference}. The framework extract features of each unlabeled data and a pool of sampled annotated images are employed to generate pseudo-labels. The pseudo-label generation process is illustrated on the right, which choose the top-k closest distances in feature space among a pool of labeled reference images and transfer their semantics.}
    \label{Reference-guided Pseudo-Label Generation}
\end{figure*}

\textbf{Label propagation} Pseudo labels can be generated indirectly through label propagation \textit{e.g.} prototype learning\cite{9757875} and nearest-neighbor matching\cite{2021Neighbor,seibold2021reference}. However, these indirect generation ways are time-consuming and demand higher memory consumption, mostly in an offline manner. For example, Han \textit{et al.} \cite{9757875} generate class representations from labeled data based on prototype learning. Through calculating the distances between feature vectors of unlabeled images and each class representation followed by a series morphological operations, high-quality pseudo labels are then generated. However, this prototype learning-based label propagation strategy requests high quality and representative feature extraction. For neighbor matching methods, Wang \textit{et al.} \cite{2021Neighbor} generate pseudo-labels on a weight basis according to the embedding similarity with neighboring labeled data. \cite{seibold2021reference} generate pseudo labels through transferring semantics that have a best fit with the unlabeled data in feature space among a pool of labeled reference images, as shown in Figure \ref{Reference-guided Pseudo-Label Generation}. Compared with network prediction-based pseudo label generation methods, label propagation-based pseudo label generation can avoid confirmation bias. Confirmation bias, which refers to the tendency a model to favor information that confirms its existing assumptions, while disregarding information that contradicts them, can be caused by the unbalanced training data and usually exists in network prediction-based pseudo label generation methods. In conclusion, these label propagation methods can premeditate the relations among data points with labeled dataset.  

Along with adding more high-confidence pseudo labels, pseudo labeling encourages low-density separation between classes. The quality of pseudo labels is the main constraint for pseudo labeling strategy. A model is unable to correct its mistakes when it overfits to a small labeled data and has confirmation bias. Then wrong predictions can be quickly amplified resulting in confident but
erroneous pseudo labels during the training process \cite{Ouali2020AnOO}. Thus, how to choose pseudo labels that will be added in the next training process and how many iterations to repeat need to be further considered. 
 
\subsection{Semi-Supervised Medical Image Segmentation with Unsupervised Regularization}

Different from generating pseudo labels and updating the segmentation model in an iterative manner, some recent progress in semi-supervised medical image segmentation has been focused on incorporating unlabeled data into the training procedure with unsupervised regularization like unsupervised loss functions. 
Algorithm \ref{Algorithm2} presents the overall workflow of this strategy. 
Different choices of the unsupervised loss functions and regularization terms lead to different semi-supervised models.
Generally, unsupervised regularization can be formulated into three sub-categories: consistency learning, co-training and entropy minimization.

\renewcommand{\algorithmicrequire}{ \textbf{Input:}}     
\renewcommand{\algorithmicensure}{ \textbf{Output:}}    
\begin{algorithm}
	\caption{Training procedure of semi-supervised learning with unsupervised regularization.}
	\label{Algorithm2}
	\begin{algorithmic}[1]
		\REQUIRE{ $\{x^{l}, y^{l}\}$ from labeled dataset $D_{L}$, $\{x^{u}\}$ from unlabeled dataset $D_{U}$, segmentation model $\mathcal{M}$}
		\ENSURE{Trained segmentation model $\mathcal{M}$}
		\WHILE{not converge}
		\STATE Calculate supervised segmentation loss $\mathcal{L}_{sup}(\theta;\mathcal{D}_{L})$
		\STATE Calculate unsupervised loss $\mathcal{L}_{unsup}(\theta;\mathcal{D})$
		\STATE Update the segmentation model $\mathcal{M}$ with the combination of supervised loss $\mathcal{L}_{sup}$ and unsupervised loss $\mathcal{L}_{unsup}$
		\ENDWHILE
		\RETURN {Trained segmentation model $\mathcal{M}$}
	\end{algorithmic}
\end{algorithm}

\subsubsection{Unsupervised Regularization with Consistency Learning}

For unsupervised regularization, consistency learning is widely applied by enforcing an invariance of predictions of input images under different perturbations and pushing the decision boundary to low-density regions, based on the assumptions that the perturbations should not change the output of the model. The consistency between two objects can be calculated as follow:
\begin{equation}\label{consistency_loss}
    Loss = D[p(x), p^{'}(T(x)]
\end{equation}
$D$ is the similarity measure function, typically using Kullback-Leibler (KL) divergence, mean squared error(MSE), Jensen-Shannon divergence(JS) and so on. $T(\cdot)$ is augmentation that adds random perturbations on data. $p$ and $p^{'}$ represent segmentation models, and their parameters can either be shared or establish a connection through certain transformations, such as exponential moving average (EMA), or they can be independent of each other. While consistency learning methods have shown promising results in semi-supervised medical image segmentation tasks due to their simplicity, it has some limitations that need to be considered:\\
1. Sensitivity to noise: Consistency learning assumes that small perturbations in the input images should not affect the model's output. However, in practice, this assumption may not always hold true as the input data can contain noise or outliers. This can lead to the model focusing on these noisy regions during training, which may reduce its generalization capability.\\
2. Hyperparameter tuning: The performance of consistency learning methods depends on the choice of hyperparameters. Selecting appropriate hyperparameters can be challenging and may require extensive experiments, making it difficult to apply these methods in practice.\\
3. Appropriate perturbations: If the perturbations are too weak, consistency-based learning may not work, but strong perturbations may confuse the model, and lead to low performance. 

\begin{figure}[!t]
    \subfigure[]{
    \label{pi Model}
    \includegraphics[width= 0.5 \textwidth]{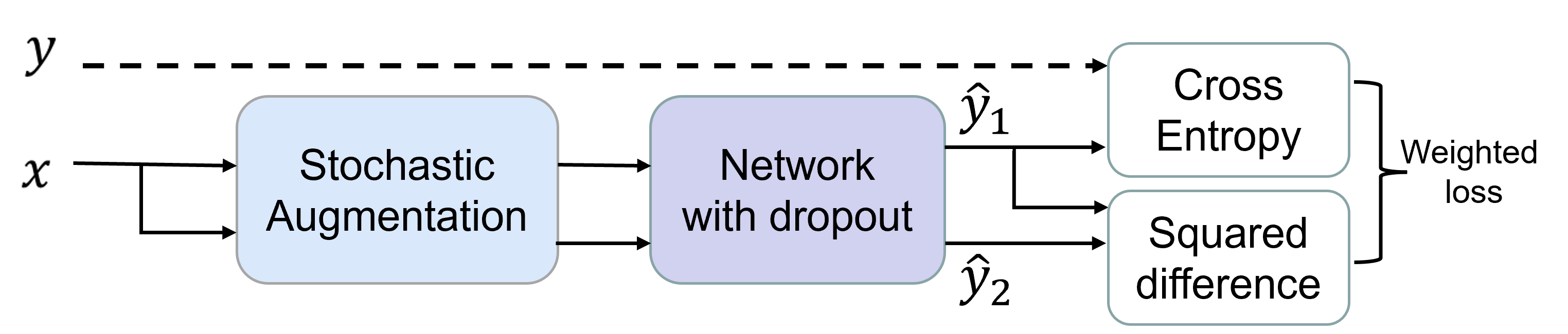}}
    \subfigure[]{
    \label{Temporal ensembling}
    \includegraphics[width= 0.5 \textwidth]{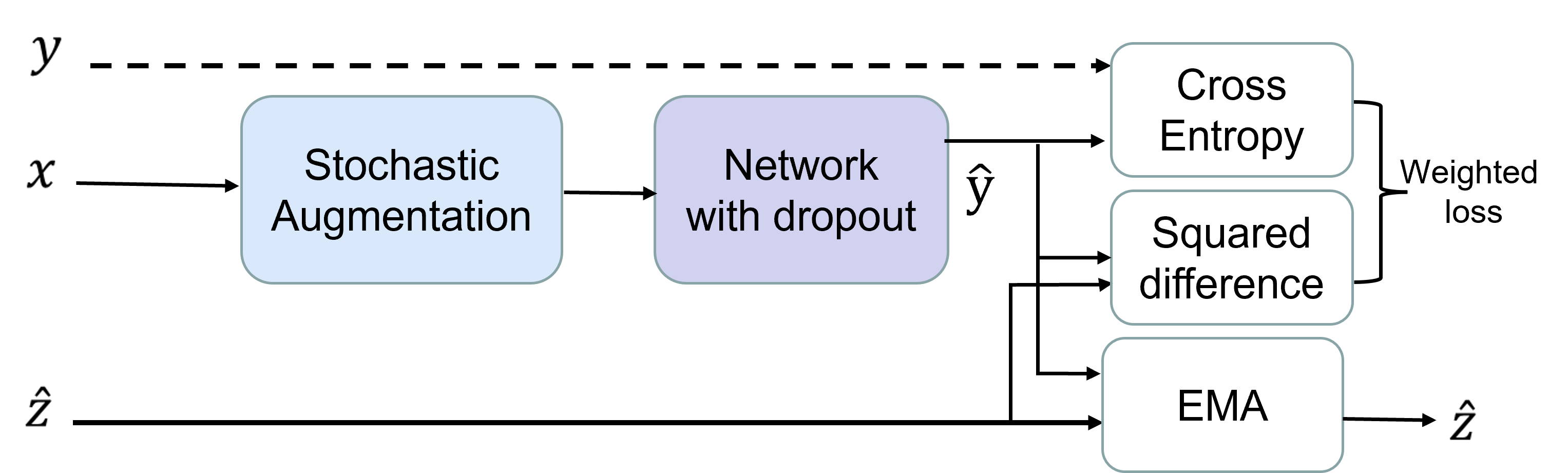}}
    \subfigure[]{
    \label{MT}
    \includegraphics[width=0.5 \textwidth]{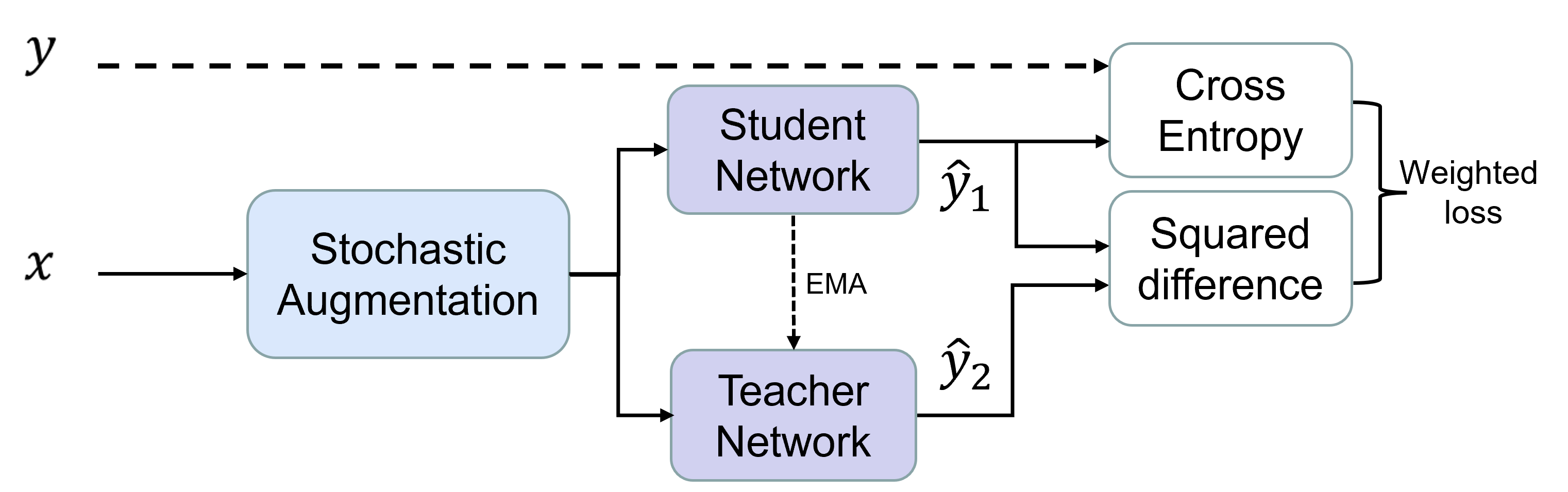}}
    \caption{The classic architectures used in consistency learning. (a): $\Pi$ Model \cite{sajjadi2016regularization} which creates two random augmentations of a sample and encourages consistent predictions. (b):Temporal ensembling strategy \cite{laine2016temporal} to use EMA predictions for unlabeled data as the consistency targets. (c):The mean-teacher architecture \cite{tarvainen2017mean}, in which the teacher model is with the EMA weights of the student model.}
    \label{The common architectures used in consistency learning}
\end{figure}

\textbf{Common architectures} The common architectures used in consistency learning in Figure \ref{The common architectures used in consistency learning} are illustrated as follows. Sajjadi \textit{et al.} \cite{sajjadi2016regularization} propose $\Pi$ Model to create two random augmentations of a sample for both labeled and unlabeled data. In the training process, the model expects the output of same unlabeled sample propagates forward twice under different random perturbations to be consistent. Samuli \textit{et al.} \cite{laine2016temporal} propose temporal ensembling strategy to use EMA predictions for unlabeled data as the consistency targets. The basic idea behind temporal ensembling is to train multiple models at different time points, and then combine their predictions to make a final prediction. However, maintaining the EMA predictions during the training process is a heavy burden. To issue the problem, Tarvainen \textit{et al.} \cite{tarvainen2017mean} propose to use a teacher model with the EMA weights of the student model for training and enforce the consistency of predictions from perturbed inputs between student and teacher models. Thus, this mean-teacher architecture is widely employed due to its simplicity. Zeng \textit{et al.} \cite{2021Reciprocal} improve the EMA weighted way in teacher models. They add a feedback signal from the performance of the student on the labeled set, through which the teacher model can be updated by gradient descent algorithm autonomously and purposefully. 

Perturbations utilized for consistency learning can be divided into input perturbations and feature map perturbations, which should be meaningful for corresponding task. The effect of perturbations on segmentation performance has an upper bound, when adding more perturbations, the segmentation performance won't be further improved \cite{9777694}. 

\textbf{Input perturbations} There are some commonly used input perturbations, such as Gaussian noise, Gaussian blurring, randomly rotation, scaling and contrast variations, and the segmentation network is encouraged to be transformation-consistent for unlabeled data \cite{li2020transformation}.
Bortsova \textit{et al.} \cite{bortsova2019semi} explore the equivariance to elastic deformations and encourage the segmentation consistency between the predictions of the two identical branches which receive differently transformed images. Huang \textit{et al.} \cite{9777694} add cutout and slice misalignment as input perturbations. Another common perturbation is mix-up augmentation \cite{berthelot2019mixmatch,9721091,2022arXiv220200677B}, which encourages the segmentation of interpolation of two data to be consistent with the interpolation of segmentation results of those data.

\begin{figure}[!t]
    \subfigure[Atrous convolution \cite{Luo2016UnderstandingTE} to enlarge foreground features]{
    \label{large}
    \includegraphics[width=0.22\textwidth]{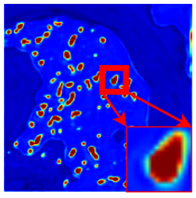}}
    \subfigure[Skip-connections \cite{he2016deep} to shrink foreground features]{
    \label{small}
    \includegraphics[width=0.22\textwidth]{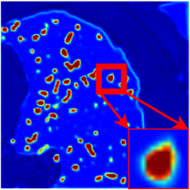}}
    \caption{Morphological feature perturbations through designing different network architectures \cite{2022arXiv220310196X}.}
    \label{Morphological Feature Perturbations through Designing Different Network Architectures}
\end{figure}

\begin{figure*}[h]
    \centering
    \subfigure[Shadow augmentation which imposes the shadow artifacts extracted from shadow source images on the original input images with different values of shadow threshold $\tau_{s}$.]{
    \label{Fig.sub.1}
    \includegraphics[scale=.5]{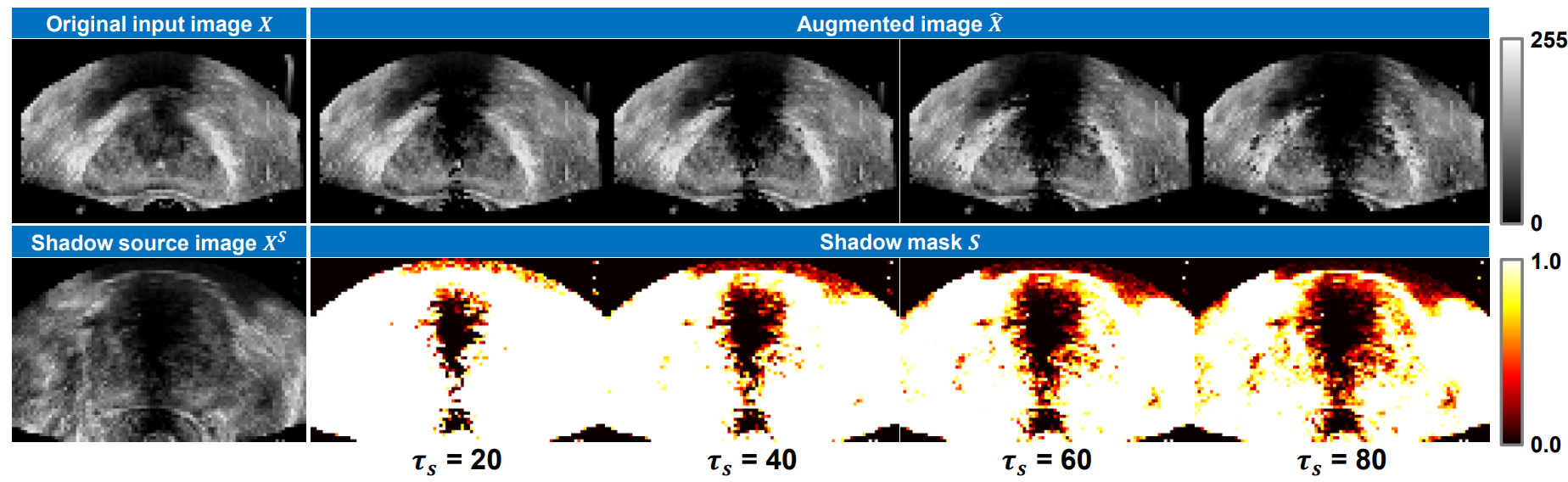}}
    \subfigure[Shadow dropout which filters the features
extracted from the shadow regions in feature maps.]{
    \label{Fig.sub.2}
    \includegraphics[scale=.5]{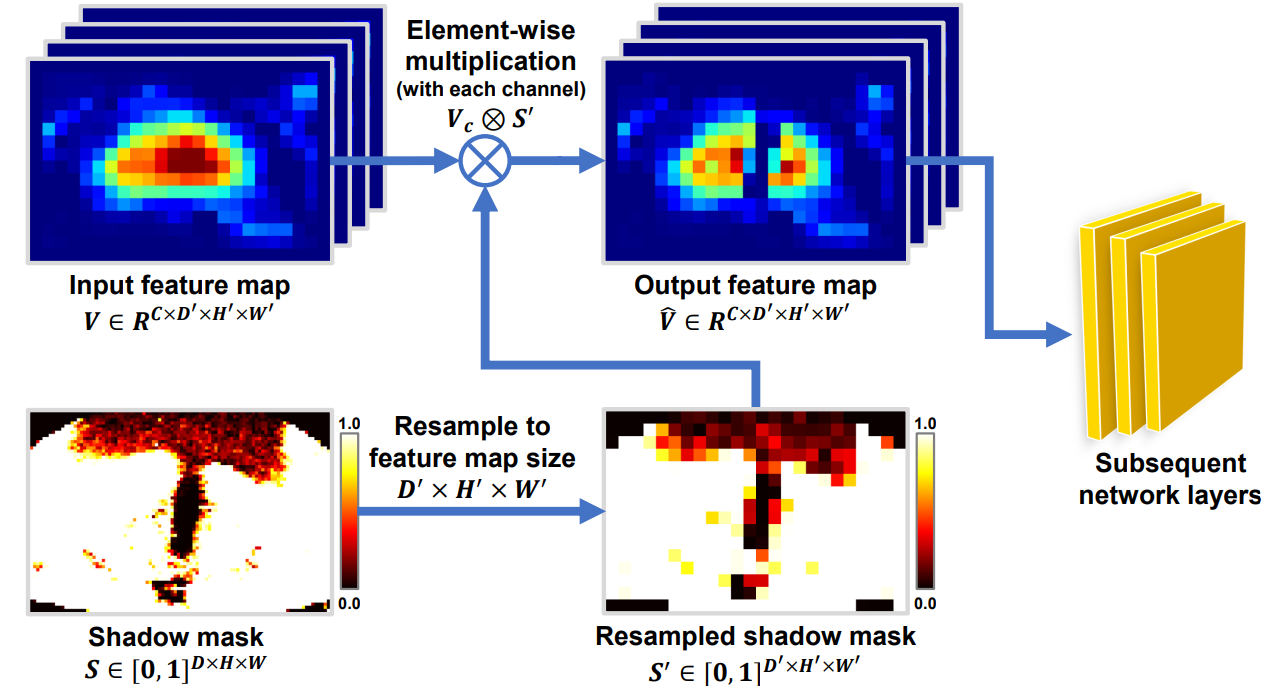}}
    \caption{Shadow augmentation and dropout \cite{xu2021shadow}}
    \label{Shadow Augmentation and Dropout}
\end{figure*}

\textbf{Feature map perturbations} Apart from disturbances on input images, there are also many studies focusing on disturbances at feature map level. Zheng \textit{et al.} \cite{zheng2022double} propose to add random noise to the parameter calculations of the teacher model. Xu \textit{et al.} \cite{2022arXiv220310196X} propose morphological feature perturbations through designing different network architectures, as shown in Figure \ref{Morphological Feature Perturbations through Designing Different Network Architectures}. Atrous convolutions can enlarge foreground features while skip-connections will shrink foreground features \cite{Luo2016UnderstandingTE,Xu2020LearningTP}. Li \textit{et al.} \cite{li2021dual} add seven types of feature perturbations to seven extra decoders and require this seven predictions to be consistent with the main decoder. These feature level perturbations are feature noise, feature dropout, object masking, context masking, guided cutout, intermediate VAT, and random dropout, based on the work in \cite{ouali2020semi}. Among them, object masking, context masking and guided cutout utilize the predictions of the decoder to mask objects or contexts in feature maps; intermediate VAT refers to using virtual adversarial training as a perturbation function for feature maps. Some studies apply perturbations both at the input and feature map levels. For example, Xu \textit{et al.} \cite{xu2021shadow} propose a novel shadow perturbation which contains shadow augmentation \ref{Fig.sub.1} and shadow dropout \ref{Fig.sub.2} to simulate the low image quality and shadow artifacts in medical images. Specifically, shadow augmentation is a perturbation through adding simulated shadow artifacts to the input images while shadow dropout will drop neural nodes according to the prior knowledge of the shadow artifacts, which is a disturbance acting directly on feature maps. However, if the perturbations are too weak, it may cause the student model to memorize these easy variations and fit the training data quickly. Finally, the student model fails to discover effective features, which is the Lazy Student Phenomenon. But strong perturbations may confuse the teacher and student, and lead to low performance. To avoid the large gap between the student model and teacher model, Shu \textit{et al.} \cite{9721091} add a transductive monitor for further knowledge distillation to narrow the semantic gap between the student model and teacher model. Some works \cite{NEURIPS2020_06964dce,kim2022conmatch,yang2023revisiting} explicitly divide perturbations into strong and weak perturbations, and use the prediction from a weakly perturbed input to supervise the prediction from its strong perturbed version. These works hold the assumption that weakly perturbed inputs can provide reliable predictions whereas strongly perturbed ones can improve the learning process and model robustness.     

\begin{figure}[!t]
    \centering
    \includegraphics[width=\linewidth]{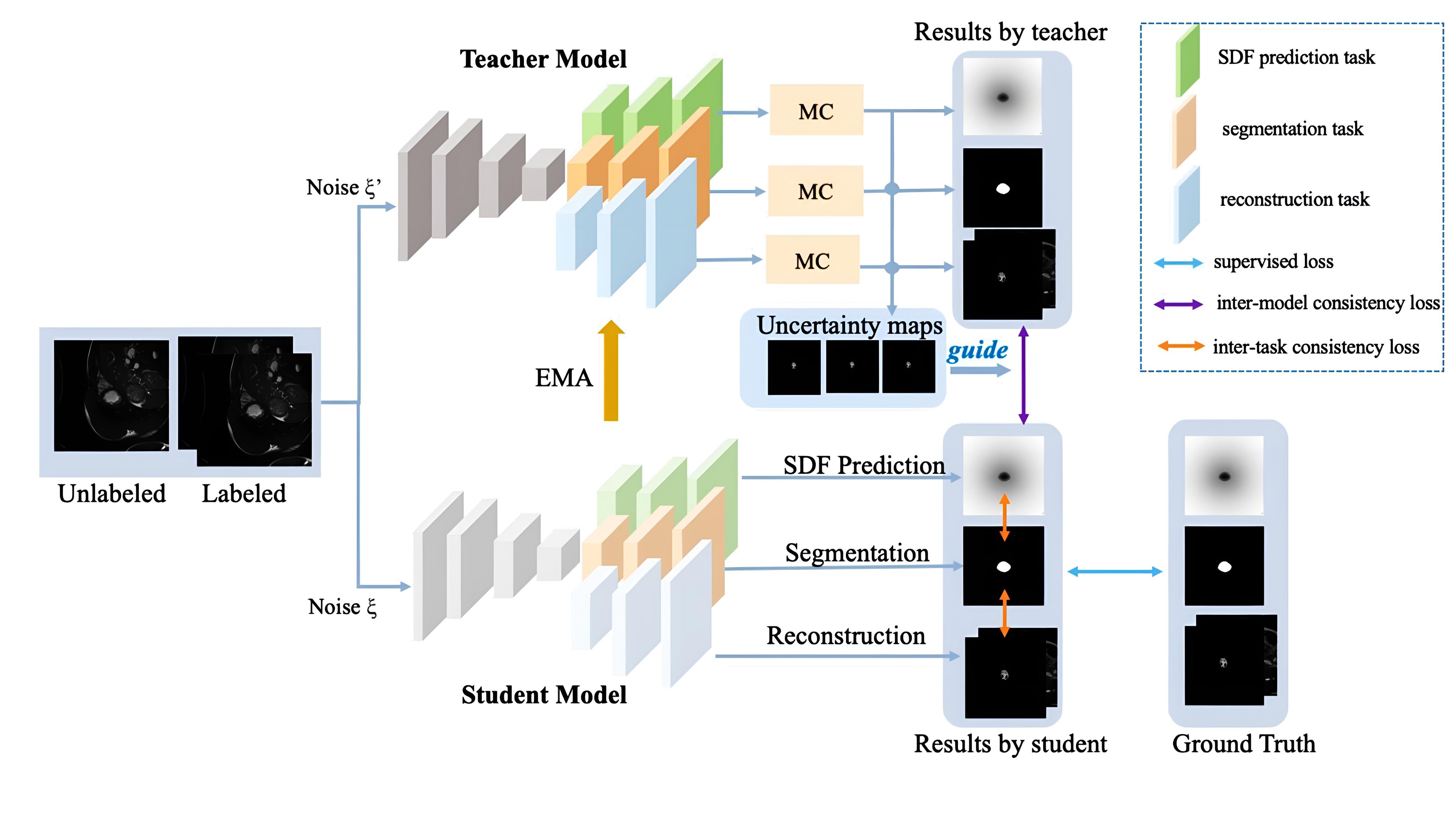}
    \caption{Multi-task learning in mean teacher architecture which contains the segmentation task, the reconstruction task, and the signed distance field prediction task\cite{wang2021tripled}. The inter-task consistency encourages consistent predictions between the three tasks and the inter-model consistency encourages consistent predictions between the teacher model and the student model.}
    \label{tripled-uncertainty guided mean teacher model}
\end{figure}

\textbf{Task-level regularization} Other than utilizing data-level perturbations for consistency learning, some methods focus on building task-level regularization by adding auxiliary task to leverage geometric information. 
Li \textit{et al.} \cite{li2020shape} develop a multi-task network to build shape-aware constraints with adversarial regularization. Liu \textit{et al.} \cite{liu2021shape} propose a shape-aware multi-task framework which contained segmentation, signed distance map prediction and organ contour prediction. Luo \textit{et al.} \cite{luo2021semi} combine the level set function regression task with the segmentation task to form a dual-task consistency for semi-supervised learning. Zhang \textit{et al.} \cite{zhang2021dual} propose dual-task mutual learning framework by encouraging dual-task networks to explore useful knowledge from each other. Based on dual-task framework, Zhang \textit{et al.} \cite{zhang2023uncertainty} utilize both segmentation task and regression task for self-ensembling and utilize estimated uncertainty to guide the mutual consistency learning and obtain further performance improvement, Shi \textit{et al.} \cite{shi2023competitive} propose to utilize segmentation task and regression task as student networks for competitive ensembling to the teacher network. Chen \textit{et al.} \cite{9689970} propose a dual-task consistency joint learning framework that encouraged the segmentation results to be consistent with the transformation of the signed distance map predictions. Wang \textit{et al.} \cite{wang2021tripled} inject multi-task learning into mean teacher architecture which contain the segmentation task, the reconstruction task, and the signed distance field prediction task so that the model can take account of the data-, model- and task-level consistency, as shown in Figure \ref{tripled-uncertainty guided mean teacher model}. In signed distance field prediction, a neural network is trained to predict the signed distance value of each pixel from the nearest foreground points. The sign of the distance indicates whether the point is inside or outside the region of interest, while the magnitude of the distance gives an estimate of the distance from the foreground. Besides, they propose an uncertainty weighted integration (UWI) strategy to estimate the uncertainty on all tasks and develop a triple-uncertainty based on these tasks to guide the student model to learn reliable information from teacher.

\textbf{Variants of consistency calculation methods} There are multiple consistency calculation methods to avoid noisy pixel predictions, such as uncertainty-based consistency learning \cite{yu2019uncertainty,xie2021semi,luo2022semi,luo2021efficient,luo2022semi,luo2021efficient,fang2022annotation}, multi-level consistency learning \cite{chen2021mtans}, attention-guided consistency learning \cite{hu2022semi} and so on. The predictions of the teacher model can be wrong at some locations and might confuse the student model in the mean-teacher architecture. So uncertainty or confidence estimation are utilized to learn from more meaningful and reliable targets during training. Yu \textit{et al.} \cite{yu2019uncertainty} extend the mean teacher paradigm with an uncertainty estimation strategy through Monte Carlo dropout \cite{kendall2017uncertainties}. To use Monte Carlo dropout in semi-supervised learning, the labeled data is used to train the model with dropout turned on. Then, the model is used to make predictions on the unlabeled data, with dropout turned on. The multiple predictions of the same unlabeled sample are then used to compute an uncertainty. This uncertainty can be used to guide the pseudo-labeling process, by identifying pixels that are likely to be mislabeled or ambiguous. Xie \textit{et al.} \cite{xie2021semi} add a confidence-aware module to learn the model confidence under the guidance of labeled data. Luo \textit{et al.} \cite{luo2022semi,luo2021efficient} calculate uncertainty using pyramid predictions in one forward pass and proposed an multi-level uncertainty rectified pyramid consistency regularization. Fang \textit{et al.} \cite{fang2022annotation} attach an error estimation network to predict the loss map of the teacher’s prediction. Then the consistency loss will be calculated on low loss pixels. Chen \textit{et al.} \cite{chen2021mtans} propose multi-level consistency loss which computes the similarities between multi-scale features in an additional discriminator, where the inputs are the segmentation regions by multiplying the unlabeled input image with predicted segmentation probability maps instead of segmentation probability maps. Hu \textit{et al.} \cite{hu2022semi} propose attention guided consistency which encourages the attention maps from the student model and the teacher model to be consistent. Zhao \textit{et al.} \cite{zhao2022cross} introduce cross-level consistency constraint which is calculated between patches and the full image.Except encouraging consistency on network segmentation results directly, generative consistency \cite{9801867} is proposed through a generation network that reconstructs medical images from its predictions of the segmentation network. Xu \textit{et al.} \cite{xu2022mt} propose contour consistency and utilize Fourier series which contained a series of harmonics as an elliptical descriptor. Through minimizing the L2 distance of the parameters between the student and the teacher branch, the model is equipped with shape awareness. However, this method needs to choose different maximum harmonic numbers for the segmentation of targets with different irregularity. Each image contains the same class object, so different images share similar semantics in the feature space. Xie \textit{et al.} \cite{9662661} introduce intra- and inter-pair consistency to augment feature maps. The pixel-level relation between a pair of images in the feature space is first calculated to obtain the attention maps that highlight the regions with the same semantics but on different images. Then multiple attention maps are taken into account to filter the low-confidence regions and then merged with the original feature map to improve its representation ability. Liu \textit{et al.} \cite{liu2022contrastive} propose contrastive consistency which encourages segmentation outputs to be consistent in class-level through foreground and background class-vectors generated from a classification network. Xu \textit{et al.} \cite{9741294} propose the cyclic prototype consistency learning (CPCL) framework which contains a labeled-to-unlabeled (L2U) prototypical forward process and an unlabeled-to-labeled (U2L) backward process. The L2U forward consistency can transfer the real label supervision signals to unlabeled data using labeled prototypes while the U2L backward consistency can directly using unlabeled prototypes to segment labeled data. 

\begin{table*}
	\caption{The summarized review of semi-supervised medical image segmentation methods with consistency learning.} \label{table_con}
	\centering
	\renewcommand\arraystretch{1.4}
        \scalebox{0.97}{
	\begin{tabular}{l|l|p{1.5cm}|p{5cm}|p{4cm}}
	    \hline
        Reference &
          2D/3D &
          Modality &
          Dataset &
          Perturbations \\
         \hline
        SSN-RCL, Huang \textit{et al.}
        \cite{9777694} &
          3D &
          Microscopy &
          Kasthuri15 \cite{kasthuri2015saturated}, CREMI \footnotemark[1]  &
          Gaussian blurring, Gaussian noise, slice misalignment, contrast variations \\
        SCO-SSL, Xu \textit{et al.}\cite{xu2021shadow} &
          3D &
          US &
          UCLA \cite{sonn2013targeted} &
          Shadow augmentation, shadow dropout \\
        SemiTC, Bortsova \textit{et al.}
        \cite{bortsova2019semi} &
          2D &
          X-Ray &
          JSRT dataset \cite{shiraishi2000development}&
          Elastic deformations \\
        GCS, Chen \textit{et al.}\cite{9801867} &
          3D &
          TOF-MRA &
          MIDAS dataset \cite{bullitt2005vessel} &
          Random perturbations \\
        DUW-SSL, Wang \textit{et al.}
        \cite{wang2020double} &
          3D &
          CT, MRI &
          LA dataset \cite{xiong2021global}, KiTS dataset \cite{heller2019kits19} &
          Random noise, dropout \\
        URPC, Luo \textit{et al.}\cite{luo2021efficient} &
          3D &
          CT &
          BraTS 2019\cite{menze2014multimodal}, Pancreas CT\cite{clark2013cancer} &
          Randomly cropped patches, multi-level pyramid predictions \\
        Mtans, Chen \textit{et al.}\cite{chen2021mtans} &
          3D &
          MRI &
          Longitudinal Multiple Sclerosis Lesion Segmentation \cite{carass2017longitudinal}, ISLES 2015\cite{maier2017isles}, BraTS 2018\cite{menze2014multimodal} &
          Multi-scale features \\
        CPCL, Xu \textit{et al.}\cite{9741294} &
          3D &
          CT, MRI &
          BraTS 2019 \cite{menze2014multimodal}, KiTS dataset \cite{heller2019kits19} &
          Different input images \\
        AHDC, Chen \textit{et al.}\cite{2021arXiv210908311C} &
          3D &
          CT, MRI &
          LGE-CMR datasets from \cite{karim2013evaluation,li2021atrialgeneral}, MM-WHS dataset \cite{zhuang2016multi,zhuang2013challenges} &
          Different domain inputs \\
        UA-MT, Yu \textit{et al.}\cite{yu2019uncertainty} &
          3D &
          MRI &
          LA dataset \cite{xiong2021global} &
          Random flipping, random rotating
           \\
        SASSNet, Zhang \textit{et al.}\cite{li2020shape} &
          3D &
          MRI &
          LA dataset \cite{xiong2021global} &
          Task-level consistency \\
        DTC, Luo \textit{et al.}\cite{luo2021semi} &
          3D &
          CT, MRI &
          LA dataset \cite{xiong2021global}, Pancreas CT \cite{clark2013cancer} &
          Task-level consistency \\
        T-UncA, Wang \textit{et al.}\cite{wang2022semi} &
          2D &
          MRI &
          ACDC dataset \cite{bernard2018deep}, PROMISE \cite{litjens2014evaluation} &
          Task-level consistency \\
		\hline
	\end{tabular}}
		\\
	\footnotetext[1]  01. https://cremi.org/
\end{table*}

\subsubsection{Unsupervised Regularization with Co-Training}
\label{sec:co-training}
Co-training framework assumes that each data has two or more different views and each view has sufficient information to give predictions independently \cite{blum1998combining}. It first learns a separate segmentation model for each view on labeled data, and then the predictions of the models on unlabeled data are gradually added to training set to continue the training. In co-training, one view is redundant to other views and the models are encouraged to have consistent predictions on all the views. Note that different from self-training methods, co-training methods add pseudo labels from one view to the training set and act as supervision signals to train models of other views. And the difference between co-training and consistency learning is that all the models in co-training will be updated through gradient descent algorithm whereas the consistency learning encourages the outputs for different perturbations to be consistent and only one main model is updated by gradient descent algorithm, such as the mean-teacher architecture \cite{tarvainen2017mean}. There are also some limitations that need to be considered:\\
1. Sufficient independence between views: Co-training assumes that each view is sufficient and independent enough to make predictions on its own. However, in real-world scenarios, this assumption may not always hold true, leading to poor performance when the views are correlated or redundant.\\
2. Risk of model conflict: Co-training encourages consistency between the models' predictions across different views. However, if the models are too similar, they may become overly specialized and fail to capture the underlying patterns in the data.\\
3. Sensitivity to noisy pseudo labels: Co-training adds pseudo labels from one view to the training set as supervision signals for other views. If these pseudo labels from one view are noisy or incorrect, it can negatively impact the performance of other views.\\

\begin{figure}[!t]
    \centering
    \includegraphics[width=0.5\textwidth]{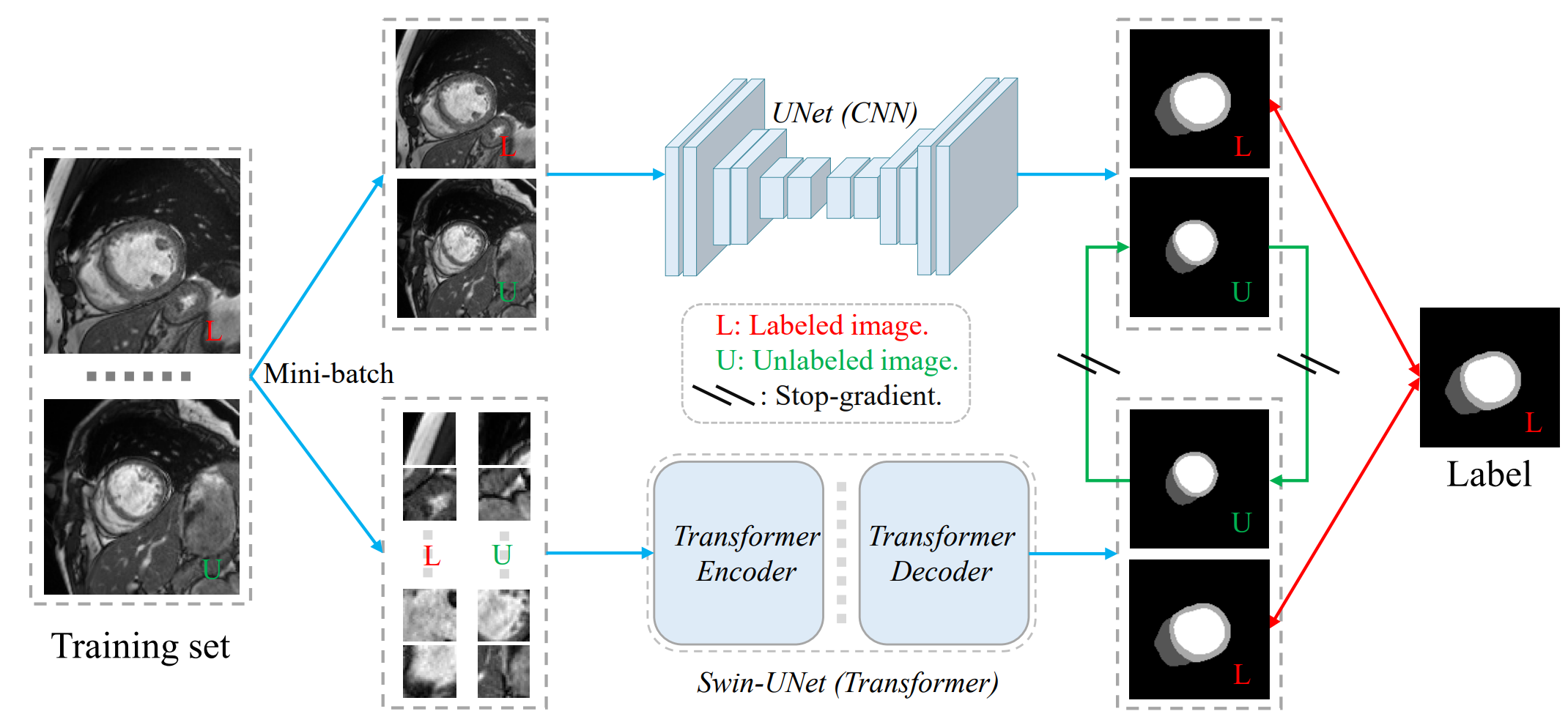}
    \caption{A co-training framework which uses CNN- and transformer-based backbones and encourages the consistency and complementary between different networks \cite{luo2022crossteaching}.}
    \label{Cross Teaching between CNN and Transformer}
\end{figure}

\textbf{Construction of different views} The core of co-training is how to construct two (or more) deep models of approximately represent sufficiently independent views. The methods mainly contain using different sources of data, employing different network architectures and using special training methods to obtain diverse deep models. First, different sources of data includes data from different modalities \cite{2021Semi,chen2022mass}, medical centers \cite{liu2022act} or anatomical planes \cite{xia20203d,zhao2022mmgl}, which lead to different distributions. For example, Zhu \textit{et al.} \cite{2021Semi} propose a co-training framework for unpaired multi-modal learning. This framework contains two segmentation networks and two image translation networks across two modalities. They utilize the pseudo-labels (from unlabeled data) or labels (from labeled data) from one modality to train the segmentation network in the other modality after image translation. For one thing, it increases supervision signals. For another, it adds modality-level consistency. Chen \textit{et al.} \cite{chen2022mass} leveraged unpaired multi-modality images to be cross-modal consistent in anatomy and semantic information. The multi modalities which are collaborative and complementary could encourage better modality-independent representation learning. Liu \textit{et al.} \cite{liu2022act}  present a co-training framework for domain-adaptive medical image segmentation. This framework contains two segmentors used for semi-supervised segmentation task (labeled and unlabeled target domain data as inputs) and unsupervised domain adaptation task (labeled source domain data and unlabeled target domain data as inputs), respectively. \cite{xia20203d,zhao2022mmgl}  use coronal, sagittal and axial views of 3D medical images as view difference at input level and \cite{xia20203d} also use asymmetric 3D kernels with 2D initialization as view difference at feature level. However, when there are only one source of data available, training two (or more) identical networks may lead to collapsed neural networks as the predictions from these models are encouraged to be similar. \cite{qiao2018deep,peng2020deep} generate adversarial examples as another view. Second, as different models usually extract different representations, different models in co-training framework can focus on different views. Except using CNN as the backbones, there are also some transformer-based backbones \cite{9761533,xiao4081789efficient}. As shown in Figure \ref{Cross Teaching between CNN and Transformer}, Luo \textit{et al.} \cite{luo2022crossteaching} introduce the cross teaching between CNN- and transformer-based backbones which implicitly encourages the consistency and complementary between different networks. Liu \textit{et al.} \cite{9761533} combine CNN blocks and Swin Transformer blocks as the backbone. Xiao \textit{et al.} \cite{xiao4081789efficient} add another teacher model with the transformer-based architecture. The teacher models communicate with each other with consistency regularization and guide the student learning process.   Third, diverse deep models can also be trained using special training methods. For instance, Chen \textit{et al.} \cite{dong2018tri} use output smearing to generate different labeled data sets to initialize diverse models. To maintain the diversity in the subsequent training process, the modules are fine-tuned using the generated sets in specific rounds.

\textbf{Avoiding noisy pseudo labels} in co-training is also important. Although consistent predictions are encouraged across the networks, they may contain noise leading to unstable training process. To overcome the third limitation mentioned in the Sec.\ref{sec:co-training}, an uncertainty-aware co-training framework \cite{xia20203d} is proposed through estimating the confidence of each view and fusing the predictions from the other views to generate the pseudo labels for one view. Wang \textit{et al.} \cite{wang2021self} develop a self-paced and self-consistent co-training framework. The self-paced strategy can encourage the network to transfer the knowledge of easier-to-segment regions to the harder ones gradually through minimizing a generalized Jensen-Shannon divergence. Another way to alleviate the influence from noisy pseudo labels is through exponential mix-up decay to adjust the contribution of the supervision signals from both labels and pseudo labels across the training process \cite{liu2022act}. Except the methods mentioned above, adversarial learning in \ref{Adversarial Learning} is always conducted to generate pixel-wise confidence maps or uncertainty. The semi-supervised models will learn from high-confidence predictions \cite{Peiris2023}, thus avoiding noisy pseudo labels.   


\begin{table*}
	\caption{The summarized review of semi-supervised medical image segmentation methods with co-training.} \label{table_co-training}
	\centering
	\renewcommand\arraystretch{1.4}
        \scalebox{0.9}{
	\begin{tabular}{l|l|l|p{6cm}|l}
		\hline 	
        Reference &
          2D/3D &
          Modality &
          Dataset &
          Diverse views from \\
         \hline
        Spsco-Cot, Wang \textit{et al.}\cite{wang2021self} &
          2D &
          CT, MRI &
          ACDC dataset \cite{bernard2018deep}, Spleen sub-task of Medical Segmentation Decathlon \cite{simpson2019large}, PROMISE \cite{litjens2014evaluation} &
          Perturbations \\
        DCT-Seg, Peng \textit{et al.}\cite{peng2020deep} &
          2D &
          CT, MRI &
          ACDC dataset\cite{bernard2018deep}, SCGM \cite{prados2017spinal}, Spleen dataset \cite{simpson2019large} &
          Perturbations \\
        MASS, Chen \textit{et al.}\cite{chen2022mass} &
          3D &
          CT, MRI &
          BTCV\cite{landman2015miccai}, CHAOS\cite{kavur2021chaos} &
          Different modalities \\
        SSUML, Zhu \textit{et al.}\cite{2021Semi} &
          2D &
          CT, MRI &
          Cardiac substructure segmentation \cite{zhuang2016multi}, Abdominal multi-organ segmentation \cite{landman2015multi,kavur2021chaos} &
          Different modalities \\
        CT\_CNN\&Trans, Luo \textit{et al.}\cite{luo2022crossteaching} &
          2D &
          MRI &
          ACDC dataset \cite{bernard2018deep} &
          Different segmentation networks \\
        Mmgl, Zhao \textit{et al.}\cite{zhao2022mmgl} &
          3D &
          CT &
          MM-WHS dataset \cite{zhuang2016multi,zhuang2013challenges} &
          Different transformations \\
        UMCT, Xia \textit{et al.}\cite{xia20203d} &
          3D &
          CT &
           NIH Pancreas \cite{clark2013cancer}, LiTS dataset \cite{bilic2019liver} & 
          Different transformations \\
		\hline
	\end{tabular}}
\end{table*}

\subsubsection{Unsupervised Regularization with Adversarial Learning}
\label{Adversarial Learning}
Adversarial methods is used to encourage the distribution of predictions from unlabeled images to be closer to that of labeled images in semi-supervised learning. These methods always contain a discriminator to distinguish the inputs from labeled annotations or unlabeled predictions \cite{zhang2017deep,yang2021medical,chen2021mtans,2021Duo}. However, adversarial training may be challenging in terms of convergence.

Zhang \textit{et al.} \cite{zhang2017deep} introduce adversarial learning to encourage the segmentations of unlabeled data to be similar with the annotations of labeled data. Chen \textit{et al.} \cite{chen2021mtans} add a discriminator following the segmentation network which is used to distinguish between the input signed distance maps from labeled images or unlabeled images. Peiris \textit{et al.} \cite{2021Duo,Peiris2023} add a critic network into the segmentation architecture which can perform the min-max game through discriminating between prediction masks and the ground truth masks. The experiments show that it could sharpen boundaries in prediction masks. The discriminator can also be used to generate pixel-wise confidence maps and select the trustworthy pixel predictions used for co-training. Wu \textit{et al.} \cite{wu2021collaborative} add two discriminators for predicting confidence maps and distinguishing the segmentation results from labeled or unlabeled data. Through adding another auxiliary discriminator, the under trained primary discriminator due to limited labeled images can be alleviated. 


\subsubsection{Unsupervised Regularization with Entropy Minimization}

Entropy minimization encourages the model to output low-entropy predictions on unlabeled data and avoids the class overlap. Semi-supervised learning algorithms \cite{wu2022cross,grandvalet2004semi,wu2021semi_EM} are usually combined with entropy minimization based on the assumption that the decision boundary should lie in low-density regions. For instance, in \cite{grandvalet2004semi}, a loss term is added to minimize the entropy of the predictions of the model on unlabeled data and the objective function turns to be as follow: 

\begin{equation}\label{maximizer of the posterior distribution}
    \begin{split}
        C(\theta,\lambda;\mathcal{L}_{n}) & = L(\theta;\mathcal{L}_{n}) - \lambda H_{emp}(Y|X,Z;\mathcal{L}_{n})\\
                & = \sum_{i=1}^n \log (\sum_{k=1}^K z_{ik} f_{k}(x_{i})) \\
                & + \lambda \sum_{i=1}^n \sum_{k=1}^K g_{k}(x_{i},z_{i}) \log g_{k}(x_{i},z_{i})
    \end{split}
\end{equation}
where $L(\theta;\mathcal{L}_{n})$ is the conditional log-likelihood and sensitive to the labeled data and $H_{emp}(Y|X,Z;\mathcal{L}_{n})$ is conditional entropy and only affected by the unlabeled data which works to minimize the class overlap. $x_{i}$ and $z_{i}$ represent inputs and corresponding labels. If $x_{i}$ is labeled $\omega_{k}$, then $z_{ik} = 1$ and $z_{il} = 0$ for $l \neq k$ ; if $X_{i}$ is unlabeled, then $z_{il} = 1$ for $l=1...k$. $f_{k}(x_{i})$ and $g_{k}(x_{i},z_{i})$ denote the model of $P(\omega_{k}|x_{i})$ and the model of $P(\omega_{k}|x_{i}, z_{i})$.  Wu \textit{et al.} \cite{wu2022cross} add entropy minimization technique in the student branch. Berthelot \textit{et al.} \cite{berthelot2019mixmatch} propose MixMatch to use a sharpening function on the target distribution of unlabeled data to minimize the entropy. The sharpening through adjusting the “temperature” of this categorical distribution is as follow:
\begin{equation}\label{sharpening}
    \begin{split}
        Sharpen(p,T)_{i} = p_{i}^{\frac{1}{T}}/\sum_{j=1}^L p_{j}^{\frac{1}{T}}
    \end{split}
\end{equation}
where $p$ is input categorical distribution and $T$ is a hyperparameter. As $T\rightarrow{}$ 0 , the output of $Sharpen(p,T)$ will approach a Dirac (“one-hot”) distribution. Lowering temperature encourages model to produce lower-entropy predictions. However, the hyperparameter needs to be set carefully and different samples may have different $T$, so \cite{wu2021semi} propose an adaptive sharpening which can adjust T adaptively for each sample according to its uncertainty predicted by the model. \cite{sajjadi2016mutual} introduce a mutual exclusivity loss for multi-class problems that explicitly forces the predictions to be mutually exclusive and encourages the decision boundary to lie on the low density space between the manifolds corresponding to different classes of data, which has a better performance in object detection task compared with  entropy minimization in \cite{grandvalet2004semi}.

Another application of entropy minimization is the use of hard label in the pseudo labeling. As argmax operation applied to a probability distribution can produce a valid “one-hot” low-entropy (i.e., high-confidence) distribution, both the entropy minimization and pseudo labeling encourages the decision boundary passing low-density regions. Therefore, the strategy of using hard label in the pseudo labeling is closely related with entropy minimization \cite{sohn2020fixmatch}. However, a high capacity model that tends to overfit quickly can give high-confidence predictions which also have low entropy \cite{oliver2018realistic}. Therefore, entropy minimization doesn't work in some cases \cite{grandvalet2004semi}.

\begin{table*}
	\caption{The summarized review of semi-supervised medical image segmentation methods with adversarial learning and entropy minimization.}
	\centering
	\renewcommand\arraystretch{1.4}
        \scalebox{0.9}{
	\begin{tabular}{c|c|c|p{2.8cm}|p{4cm}|c}
	    \hline
        Reference &
          2D/3D &
          Modality &
          Dataset &
          Highlights &
          Class \\
        \hline
        CAFD, Wu \textit{et al.}\cite{wu2021collaborative} &
          2D &
          Colonoscope &
          Kvasir-SEG \cite{jha2020kvasir}, CVC-Clinic DB \cite{bernal2015wm} &
          Introduce collaborative and adversarial learning of focused and dispersive representations &
          Adversarial learning \\
        SSTD-Aug, \textit{Chaitanya et al.} \cite{chaitanya2019semi} &
          2D &
          MRI &
          ACDC dataset \cite{bernard2018deep} &
          Task-driven data augmentation method to synthesize new training examples &
          Adversarial learning \\
        DAN, Zhang \textit{et al.}\cite{zhang2017deep} &
          3D &
          Microscopy &
          Gland Segmentation Challenge dataset \cite{sirinukunwattana2017gland} &
          Introduce adversarial learning to encourage the segmentation output of unlabeled data to be similar with the annotations of labeled data. &
          Adversarial learning \\
        GAVA, Li \textit{et al.}\cite{li2021semi} &
          2D &
          MRI &
          M\&Ms  dataset \cite{campello2021multi} &
          Employ U-net as encoder and conditional GAN as decoder &
          Adversarial learning \\
        LeakGAN\_ssl, Hou \textit{et al.}\cite{hou2022semi} &
          2D &
          Fundus &
          DRIVE\cite{staal2004ridge}, STARE\cite{hoover2000locating}, CHASE\_DB1\cite{fraz2012ensemble} &
          Add a leaking GAN to pollute the discriminator by leaking information from the generator for more moderate generations &
          Adversarial learning \\
        LG-ER-MT, Hang \textit{et al.}\cite{hang2020local} &
          3D &
          MRI &
          LA dataset \cite{xiong2021global} &
          Introduce the entropy minimization principle to the student network &
          Entropy minimization \\
        MC-Net, Wu \textit{et al.}\cite{wu2021semi} &
          3D &
          MRI &
          LA dataset \cite{xiong2021global} &
          Adjust sharpening temperature adaptively according to the uncertainty predicted by the model &
          Entropy minimization \\
		\hline
	\end{tabular}}
\end{table*}

\subsection{Semi-Supervised Medical Image Segmentation with Knowledge Priors}

Knowledge priors are the information that a learner already has before it learns new information, and sometimes are helpful for dealing with new tasks. Compared with non-medical images, medical images have many anatomical priors such as the shape and position of organs and incorporating the anatomical prior knowledge in deep learning can improve the performance for medical image segmentation \cite{zheng2019semi}. Some semi-supervised algorithms utilize knowledge priors to improve the representation ability for new tasks. While knowledge priors can be helpful in semi-supervised medical image segmentation, there are also several limitations to consider:\\
1. Overfitting: If the prior knowledge is too specific to the training data, it may lead to overfitting, where the model performs well on the training data but poorly on new, unseen data.\\
2. Non-differentiable: Some complex priors, such as region connectivity, convexity and symmetry are usually non-differentiable and complex losses need to be designed.  
In this part, we categorize the knowledge priors as self-supervised tasks and anatomical priors.

\begin{figure}[!t]
    \centering
    \includegraphics[width=0.5\textwidth]{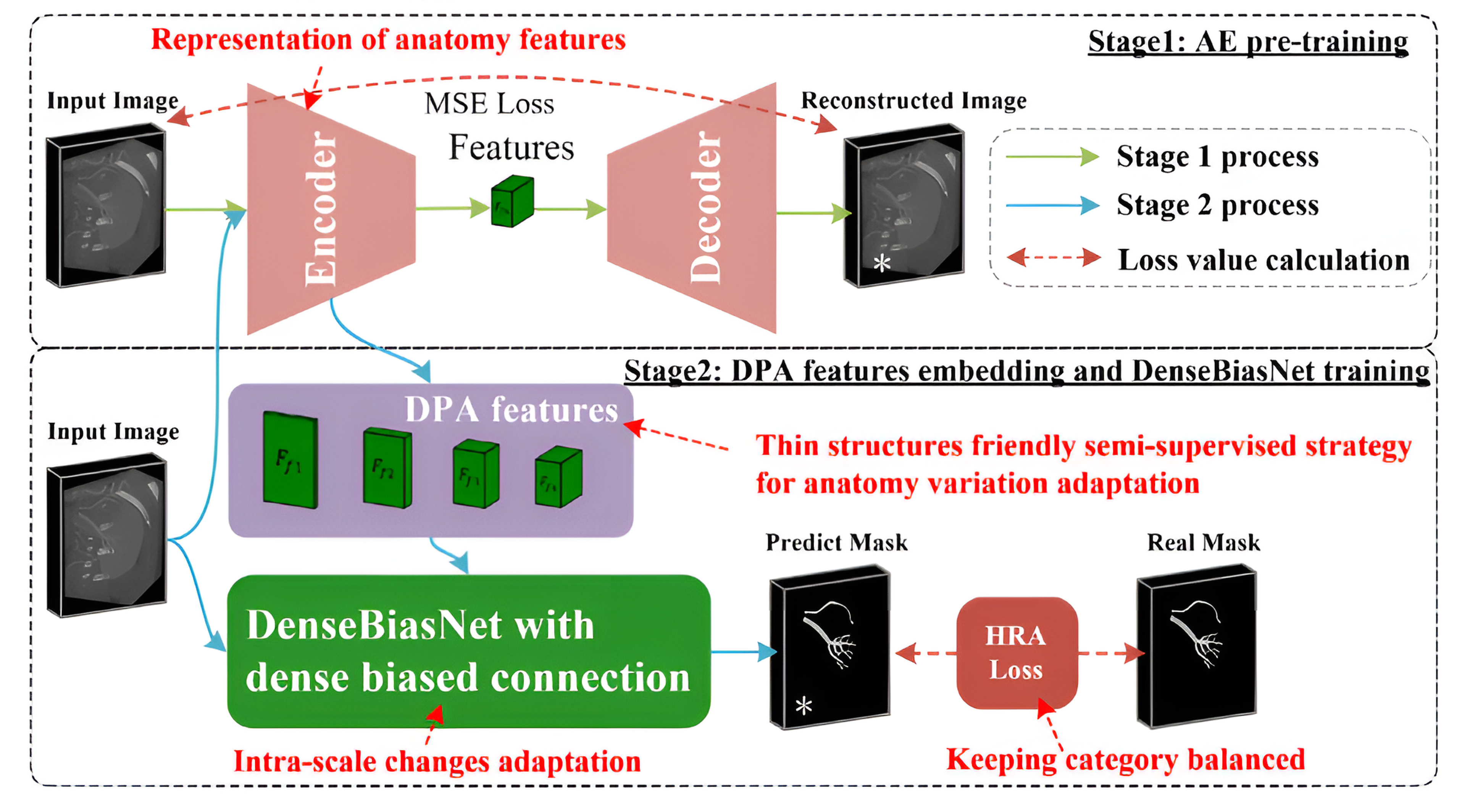}
    \caption{DPA-DenseBiasNet \cite{he2020dense} for fine renal artery segmentation. An auto-encoder of a reconstruction task is trained in stage 1 process. The deep prior anatomy (DPA) features extracted from the encoder, which contain representations of anatomy priors,  are then embedded for the downstream segmentation task in stage 2 process.}
    \label{DPA-DenseBiasNet}
\end{figure}

Self-supervised tasks which employs large unlabeled data to train networks can provide useful representations and visual priors. An important role is to pretrain networks and provide better starting points for target tasks. For example, huang \textit{et al.} \cite{9777694} add a reconstruction pre-training from the counterparts to avoid networks being randomly initialized in a cold start stage. Wang \textit{et al.} \cite{wang2022separated} use superpixels to separate an image into regions and learned intra- and inter-organ representation based on contrastive learning, then the model is used to initialize the semi-supervised framework, which boost the performance significantly. Self-supervised tasks can also be trained jointly with target semi-supervised tasks as regularization. Contrastive learning are the most popular methods to integrate with semi-supervised framework. For example,  hu \textit{et al.} \cite{hu2021semi} introduce the self-supervised image-level and pixel-level contrastive learning into the semi-supervised framework. \cite{peng2021self} integrate self-paced contrastive learning. Wu \textit{et al.} \cite{wu2022cross} add patch- and pixel-level dense contrastive loss to align the features from the teacher and student models. Zhao \textit{et al.} \cite{zhao2022mmgl} introduce the multi-scale multi-view global-local contrastive learning into co-training framework. However, in contrastive learning, negative samples may come from the similar features from anchors, which may confuse models during training. You \textit{et al.} \cite{you2023rethinking} integrate contrastive learning from a variance-reduction perspective, which uses stratified group sampling theory and generalize well in long-tail distribution. Except contrastive learning, jigsaw puzzle tasks \cite{9756342}, lesion region inpainting \cite{zhang2021self} and reconstruction tasks \cite{he2020dense} can also be utilized easily into semi-supervised framework. \cite{zhang2021self} propose a dual-task network with a shared encoder and two independent decoders for lesion region inpainting and segmentation. They also add entropy minimization technique in the student branch. He \textit{et al.} \cite{he2020dense} train an auto-encoder through a reconstruction task and the deep prior anatomy (DPA) features extracted from it are then embedded for segmenting, as shown in Figure \ref{DPA-DenseBiasNet}.

\begin{figure*}[!t]
    \includegraphics[width=\textwidth]{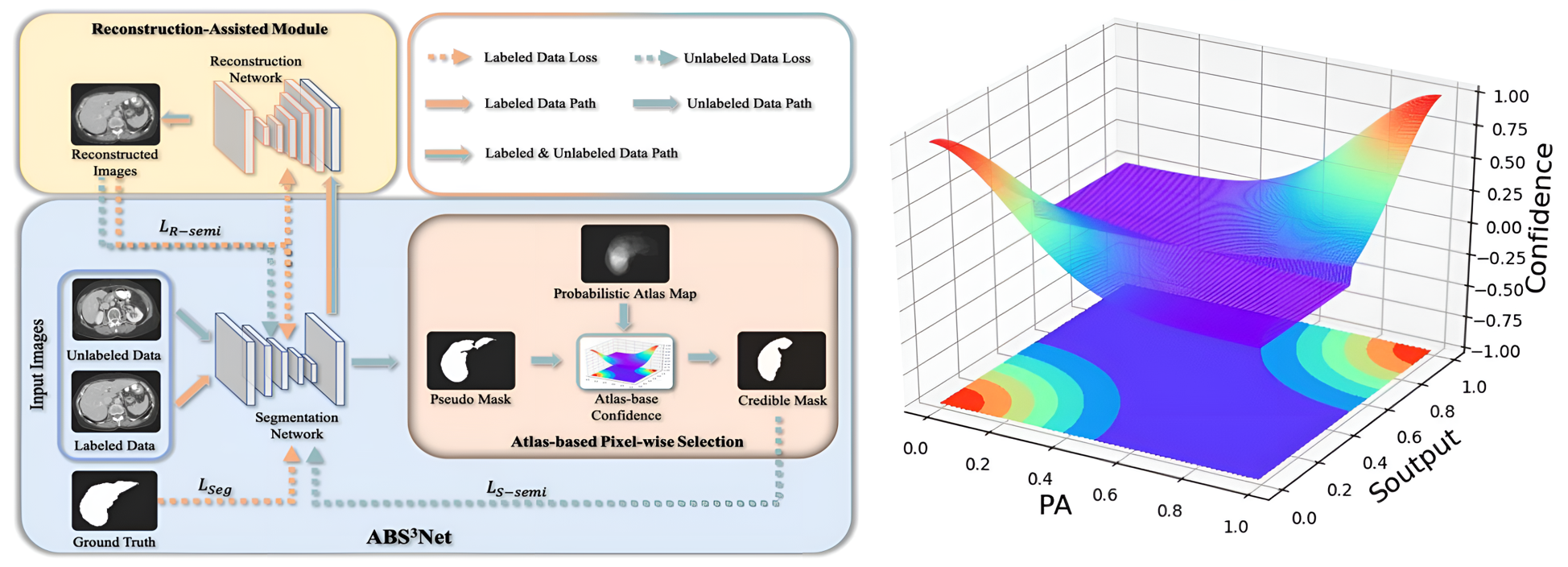}
    \caption{Illustration of the framework of $ABS^{3}Net$ \cite{9721677} with the confidence map. The left is the $ABS^{3}Net$, in which atlas-based pixel selection module is introduced to select reliable pixel results based on pixel-wise confidence. The right shows the atlas-based confidence map. The high confidence (shown in red) represents both probability atlas map (PA) and the segmentation probability map $s_{output}$ are close to the prediction mask $s_{mask}$. The confidence decreases from red to blue.}
    \label{confidence atlas}
\end{figure*}

\begin{figure}[!t]
    \subfigure[]{
    \label{Fig.si}
    \includegraphics[width=0.15\textwidth]{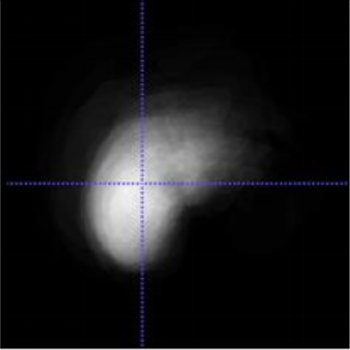}}
    \subfigure[]{
    \label{Fig.lr}
    \includegraphics[width=0.15\textwidth]{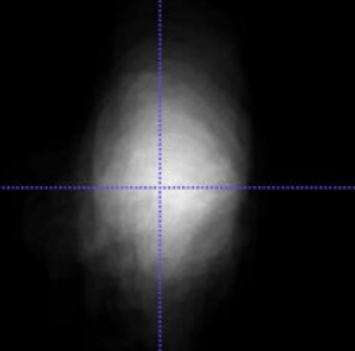}}
    \subfigure[]{
    \label{Fig.ap}
    \includegraphics[width=0.15\textwidth]{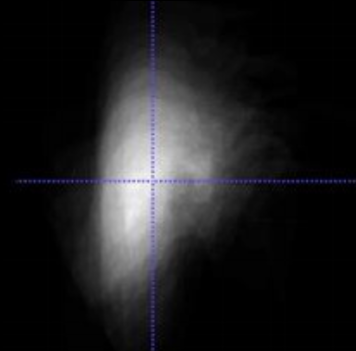}}
    \caption{The 3D probabilistic atlas of liver organ \cite{9721677}, which indicates the probability of liver pixels appearing at some location. (a)-(c) are superior–inferior, left–right direction and anterior–posterior direction correspondingly.}
    \label{The 3D Probabilistic atlas of liver organ}
\end{figure}

Anatomical priors include fixed locations, shapes, region sizes and anatomical relations and so on. Objects, such as organs, in medical segmentation usually have fixed locations and shapes. To take position information and shape prior into account, atlas maps are widely applied in medical image segmentation\cite{dong2015segmentation,park2003construction,zheng2019semi,9721677,li2022coupling}. As shown in Figure \ref{The 3D Probabilistic atlas of liver organ}, an atlas map, which indicates the probability of objects appearing at some location, can be generated as follows. First, annotated volumes need to be registered to a referenced volume. Then the probabilistic atlas (PA) can be generated through averaging manually masks after deformable of all annotated volumes. For example, Zheng \textit{et al.} \cite{zheng2019semi} calculate the liver PA and predefined the hard pixel samples with the atlas values close to 0.5. Huang \textit{et al.} \cite{9721677} utilize PA to give the unlabeled data segmentation pixel-wise confidence to select reliable pixel results, as shown in Figure \ref{confidence atlas}. The pixel-wise confidence is calculated as follows: 
\begin{equation}\label{atlas confidence}
    \begin{split}
        Confidence = \exp(-\frac{(PA-s_{mask})^2 + (s_{output}-s_{mask})^2}{2\sigma ^2})
    \end{split}
\end{equation}
\begin{equation}\label{s_mask}
    \begin{split}
        s_{mask} = [s_{output}+0.5]
    \end{split}
\end{equation}
where $s_{output}$ and $s_{mask}$ refer to the segmentation probability map for the organ to be segmented and the prediction mask of unlabeled data, whose value is only 0 or 1. $[\cdot]$ denotes the integer-valued function. As can be seen in Figure \ref{confidence atlas}, the confidence decreases from red to blue and the confidence is higher, when both PA and $s_{output}$ are close to $s_{mask}$, that is 0 or 1. However, segmentation algorithms utilizing atlas maps may be unsuitable for targets that have large positional variance. Furthermore, the segmentation performance highly relies on accurate registration. Fixed locations and shapes are always utilized in organ segmentation whereas anatomical relations can be utilized in multi-type pathology segmentation. Anatomical relations represent relative locations of different objects. For example, MyoPS-Net \cite{qiu2023myops} uses inclusiveness loss to represent relations between different types of pathologies, which constrains the pixels of scars to be included in the pixels of edema. \cite{Chen2022MagicNetSM} propose magic-cube partition and recovery, encouraging unlabeled images to learn organ semantics in relative locations from labeled images. The limitation of this magic-cube partition and recovery augmentation is that it may not work well on unaligned images. Another assumption is that objects from the same class across all the samples share the same anatomical adjacencies, despite their varying region geometries, thus an adjacency-graph based auxiliary training loss that penalizes outputs with anatomically incorrect region relationships is introduced in \cite{ganaye2019removing}. For size priors, PaNN\cite{zhou2019prior} constrains the predicted average distribution of organ sizes to be similar with the prior statistics from the labeled dataset. 

The algorithms mentioned above are typically simple whereas some complex priors, such as region connectivity, convexity, symmetry are usually non-differentiable. Therefore, specific losses need to be designed for these complex constraints. In \cite{wang2023cat}, an out-of-box and differentiable way to consider complex anatomical priors is developed based on reinforce algorithm and adversarial samples. Experiments show that clinical-plausible segmentations are obtained. Another work in \cite{clough2020topological} introduces persistent homology, a concept from topological data analysis, to specify the desired topology of segmented objects in terms of their Betti numbers and then drive the predictions of unlabeled data to contain the specified topological features. The Betti numbers count the number of features of some dimension, such as the number of connected components, the number of loops or holes, the number of hollow voids and so on. This process does not require any ground-truth labels, just prior knowledge of the topology of the structure being segmented. The idea of persistent homology can be applicable for segmentation of objects with a fixed and regular shape, such as cardiac chambers and myocardium.    

Rich knowledge priors make medical image segmentation different from natural image segmentation. In semi-supervised medical image segmentation with limited labeled data, more accurate and plausible results can be obtained by incorporating medical knowledge priors.

\begin{table*}
	\caption{The summarized review of semi-supervised medical image segmentation methods with knowledge priors.}
	\centering
	\renewcommand\arraystretch{1.4}
        \scalebox{0.9}{
	\begin{tabular}{l|l|p{2cm}|p{3cm}|p{4cm}|p{2cm}}
        \hline
        Reference &
          2D/3D &
          Modality &
          Dataset &
          Highlights &
          Class \\
        \hline
        SepaReg, Wang \textit{et al.}\cite{wang2022separated} &
          3D &
          CT &
          PDDCA\cite{raudaschl2017evaluation}&
          Initialization with pre-trained model based on intra- and inter-organ contrastive learning &
          Contrastive learning \\
        S4 ML, Kiyasseh \textit{et al.}\cite{kiyasseh2021segmentation} &
          2D &
          MRI &
          LA dataset \cite{xiong2021global} &
          Using dataform multi centers through meta-learning and contrastive learning task performed with unlabelled data &
          Contrastive learning \\
        Le-SSCL, Hu \textit{et al.}\cite{hu2021semi} &
          3D &
          CT, MRI &
          Hippocampus subset of Medical Segmentation Decathlon \cite{simpson2019large}, MM-WHS dataset \cite{zhuang2016multi,zhuang2013challenges} &
          Self-supervised image-level and supervised pixel-level contrastive pre-training &
          Contrastive learning \\
        SimCVD, You \textit{et al.} \cite{9740182} &
          3D &
          CT, MRI &
          LA dataset \cite{xiong2021global}, Pancreas CT \cite{clark2013cancer} &
          Contrastive distillation of voxel-wise representation with signed distance maps &
          Contrastive learning \\
        CPDC, Wu \textit{et al.}\cite{wu2022cross} &
          2D &
          Microscope &
          DSB\cite{caicedo2019nucleus} , MoNuSeg\cite{kumar2019multi} &
          Cross-patch dense contrastive learning framework &
          Contrastive learning \\
        Dt-DDCL, Zhang \textit{et al.}\cite{zhang2021self} &
          2D &
          Colonoscope &
          kvasir-SEG dataset\cite{jha2020kvasir}, Skin lesion dataset \cite{gutman2016skin} &
          Dual-task network for segmentation and lesion region inpainting. &
          Inpainting task \\
        RLS\_SSL, Yang \textit{et al.} \cite{9756342} &
          3D &
          OCT &
          Private &
          Add self-supervised jigsaw puzzle task into training &
          Jigsaw puzzle task \\
        MTL-ABS3Net, Huang \textit{et al.} \cite{9721677} &
          3D &
          CT &
          LiTS dataset \cite{bilic2019liver} &
          Utilize prior anatomy to give the unlabeled data segmentation pixel-wise confidence &
          Atlas priors \\
        DAP, Zheng \textit{et al.} \cite{zheng2019semi} &
          3D &
          CT &
          LiTS dataset \cite{bilic2019liver} &
          Semi-supervised adversarial learning with deep atlas prior &
          Atlas priors \\
        \hline
	\end{tabular}}
\end{table*}

\subsection{Other Semi-Supervised Medical Image Segmentation Methods}

A frequently encountered obstacle in medical imaging is that, in real-world applications, the acquired data and annotations may be difficult to meet the assumptions, thus affecting the performance of semi-supervised learning.
Other than these methodological developments for semi-supervised segmentation methods mentioned above, We have also compiled some different concerns in real-world applications.

As there is usually a large amount of unlabeled data in semi-supervised learning, the distribution of labeled and unlabeled data may be misaligned. For better leverage of large scale data from different distributions or medical centers, some methods are proposed to deal with distribution misalignment\cite{zhang2021semi,2021arXiv210908311C,kiyasseh2021segmentation}. Zhang \textit{et al.} \cite{zhang2021semi} try to align labeled data distribution and unlabeled data distribution through minimising the L2 distance between the feature maps of them. Meanwhile, to remain discriminative for the segmentation of labeled and unlabeled data, further segmentation supervision is obtained through comparing the non-local semantic relation matrix in feature maps from the ground truth label mask and the student inputs. Another work in \cite{2021arXiv210908311C} propose adaptive hierarchical dual consistency to use the dataset from different centers, which learns mapping networks adversarially to align the distributions and extend consistency learning into intra- and inter-consistency in cross-domain segmentation. Another idea for using data from multi centers is through meta-learning. In \cite{kiyasseh2021segmentation}, one distinct task is formulated for each medical centre such that a segmentation task is performed for a centre with labelled data while the contrastive learning task is performed with unlabelled data.

Another concern in semi-supervised learning is how to fuse different supervision signals for label-efficient semi-supervised learning. As existing public imaging datasets usually have different annotations for different tasks, like CT images singly labelled tumors or partially labelled organs. Zhang \textit{et al.} \cite{zhang2021automatic} propose a dual-path semi-supervised conditional nnU-Net that can be trained on a union of partially labelled datasets, segmentation of organs at risk or tumors. Another situation is the integration of different levels of supervision signals. \cite{reiss2021every} propose multi-label deep supervision in semi-supervised framework, which leveraged image-level, box-level and pixel-level annotations. If only image-level or box-level labels  exist, the pseudo labels would be constrained to the classes contained in that or to lie within coarse regions. Except that, the noisy pseudo labels generated from the teacher model is smoothed using max-pooling to match different level predictions from the decoder for multi-level consistency. 

Class imbalance is a common problem in segmentation. In semi-supervised learning, class imbalance and limited labeled data may further bring the confirmation bias and uncertainty imbalance problem. Recently, some researchers propose class-imbalanced methods in semi-supervised learning\cite{10.1007/978-3-031-16452-1_11,wang2022ssa,wang2023dhc}. Lin \textit{et al.} \cite{10.1007/978-3-031-16452-1_11} propose a dual uncertainty-aware sampling strategy to sample low-confident categories of pixels for unsupervised consistency learning. Another direction focuses on utilizing re-weighting strategies calculated by the pixel proportion of categories\cite{wang2022ssa,wang2023dhc}.

Besides, most of previous semi-supervised frameworks are discriminative models, where labeled data is only used in the early training stage and the model may tend to overfit to the labeled data \cite{wang2022rethinking}. Wang \textit{et al.} \cite{wang2022rethinking} proposed a Bayesian deep learning framework for semi-supervised segmentation. In that way, both labeled and unlabeled data are utilized to estimate the joint distribution, which alleviates potential overfitting problem caused by using labeled data for early training only.

\section{Analysis of Empirical Results for Semi-Supervised Medical Image Segmentation}

\subsection{Common Evaluation Metrics for Medical Image Segmentation}

For medical image segmentation tasks, Dice Similarity Coefficient (DSC) is a widely used metric to measure the region overlap ratio of the ground truth $G$ and segmentation result $S$. Another similar metric IoU (or Jaccard) is used as an alternative for the evaluation. These two metrics are defined as follows: 
\begin{equation}
    DSC = \frac{2|G\cap S|}{|G|+ |S|}, \quad
    IoU = \frac{|G\cap S|}{|G\cup S|}.
\end{equation}

However, region-based metrics like DSC cannot well reflect the boundary error or small region of mis-segmentation. To issue this limitation, boundary-based evaluation metrics like Hausdorff Distance (HD) are applied to focus on the boundary distance error defined as follows:

\begin{equation}
    HD(\partial G, \partial S) = \max(\max\limits_{x\in \partial G} \min\limits_{y\in \partial S} ||x-y||_2, \max\limits_{x\in \partial S} \min\limits_{y\in \partial G} ||x-y||_2),
\end{equation}
where $\partial G$ and $\partial S$ represent the boundary of the ground truth and the segmentation result, respectively.
To eliminate the influence caused by small subsets of outliers,  95\% Hausdorff Distance (95HD) is also widely used, which is based on the calculation of the 95th percentile of the distances between boundary points.

\begin{table*}
	\caption{Representative works and empirical results on semi-supervised LA MRI segmentation benchmark.}             
        \label{la_benchmark}
	\centering
	\renewcommand\arraystretch{1.4}
        \scalebox{0.9}{
	\begin{tabular}{c|p{8.5cm}<{\centering}|c|c|c}
		\hline 	
		Method  & Highlights & $\mathcal{D}_{L}/\mathcal{D}_{U}$ & Dice & Publication\&Year \\ \hline
		\multirow{2}{*}{Baseline V-Net \cite{milletari2016v}} & \multirow{2}{*}{Fully supervised baseline with only labeled data } 
		& 8/0 & 79.99 & \\ && 16/0 & 86.03 &  \\
		Upper-bound V-Net \cite{milletari2016v} & Fully supervised upper bound with all annotations & 80/0 & 91.14 & \\ \hline
		\multirow{2}{*}{UA-MT, Yu \textit{et al.} \cite{yu2019uncertainty}} & \multirow{2}{*}{Teacher-student framework with the guidance of uncertainty}
		& 8/72 & 84.25 & \multirow{2}{*}{MICCAI 2019} \\   && 16/64 & 88.88 & \\
		\multirow{2}{*}{SASS, Li \textit{et al.} \cite{li2020shape}}  & \multirow{2}{*}{Incorporating signed distance maps for shape regularization}
		& 8/72 & 87.32  & \multirow{2}{*}{MICCAI 2020} \\  && 16/64 & 89.54 & \\
		\multirow{2}{*}{DUWM, Wang \textit{et al.}  \cite{wang2020double}}  & \multirow{2}{*}{Utilizing both segmentation and feature uncertainty}
		& 8/72 & 85.91 &  \multirow{2}{*}{MICCAI 2020} \\  && 16/64 & 89.65 &  \\ 
		\multirow{2}{*}{LG-ER-MT, Hang \textit{ et al.}   \cite{hang2020local}}  & \multirow{2}{*}{\makecell[c]{Entropy minimization to produce high-confident predictions \\ and local structural consistency to encourage \\ inter-voxel similarities }}
		& 8/72 & 85.54 &  \multirow{2}{*}{MICCAI 2020} \\  && 16/64 & 89.62 &  \\ 
		DTC, Luo  \textit{et al.}  \cite{luo2021semi}    & Encourage the consistency between output segmentation maps and signed distance map & 16/64 & 89.42 & AAAI 2021 \\
		\multirow{2}{*}{PDC-Net, Hao \textit{et al.} \cite{hao2021parameter}}  & \multirow{2}{*}{\makecell[c]{Parameter decoupling to encourage consistent predictions from \\two branch network}}
		& 8/72 & 86.55 &  \multirow{2}{*}{ICMV 2021} \\  && 16/64 & 89.76 &  \\ 
		HCR-MT, Li \textit{et al.} \cite{Li2021HierarchicalCR}    & Teacher-student framework with multi-scale deep supervision and hierarchical consistency regularization & 16/64 & 90.04 & EMBC 2021 \\
		DTML, Zhang \textit{et al.} \cite{zhang2021dual}    & Mutual learning of dual-task networks for generating segmentation and signed distance maps & 16/64 & 90.12 & PRCV 2021 \\
		\multirow{2}{*}{MC-Net, Wu \textit{et al.} \cite{wu2021semi}}  & \multirow{2}{*}{\makecell[c]{Consistency learning between outputs from \\ two different decoders}}
		& 8/72 & 87.71 &  \multirow{2}{*}{MICCAI 2021} \\  && 16/64 & 90.34 &  \\
		\multirow{2}{*}{CASS, Liu \textit{et al.} \cite{liu2022contrastive}}  & \multirow{2}{*}{Contrastive consistency on class-level}
		& 8/72 & 86.51 &  \multirow{2}{*}{CMIG 2022} \\  && 16/64 & 89.81 &  \\ 
		\multirow{2}{*}{SimCVD, You \textit{et al.} \cite{9740182}}  & \multirow{2}{*}{\makecell[c]{Contrastive distillation of voxel-wise representation with signed \\ distance maps}}
		& 8/72 & 89.03 &  \multirow{2}{*}{TMI 2022} \\  && 16/64 & 90.85 &  \\
		\multirow{2}{*}{CMM, Shu \textit{et al.} \cite{9721091}}  & \multirow{2}{*}{\makecell[c]{Asynchronously perform Cross-Mix Teaching and transductive \\ monitor for active knowledge  distillation}} & 8/72 & 85.92 & \multirow{2}{*}{TMM 2022} \\ && 16/64 & 90.03 & \\

		DTCJL, Chen \textit{et al.}  \cite{9689970}    & Semi-supervised dual-task consistent
joint learning framework with task-level regularization & 16/64 & 90.32 & TCBB 2022 \\
		\hline
	\end{tabular}}
\end{table*}

\subsection{Benchmark Datasets for Semi-Supervised Medical Image Segmentation}
\label{sec:benchmark}

In addition to the promising progress in semi-supervised medical image segmentation methods, several segmentation benchmarks are also evolved to ensure a fair comparison of these methods with the same task setting on same public dataset.

\textbf{LA dataset. }
The LA benchmark dataset \cite{xiong2021global} from the Left Atrium Segmentation Challenge \footnote{http://atriaseg2018.cardiacatlas.org/data/} contains 100 3D gadolinium-enhanced MR imaging scans (GE-MRIs) for training, with an isotropic resolution of $0.625 \times 0.625 \times 0.625 mm^{3}$ .
Since the testing set of LA does not include public annotations, for the settings in \cite{yu2019uncertainty}, the 100 training scans are splitted into 80 scans for training and 20 scans for testing. Out of the 80 training scans, 20\% (i.e. 16 scans) are used as labeled data and the remaining as unlabeled data. V-Net \cite{milletari2016v} is used as the network backbone for all experiments with a joint cross-entropy loss and dice loss for training. For supervised comparisons, V-Net trained with only labeled data (i.e. 16 scans) and trained with all labeled data (i.e. 80 scans) is performed as lower bound and upper bound for semi-supervised learning.
As shown in Table. \ref{la_benchmark}, as one of the most popular benchmark dataset for semi-supervised medical image segmentation, many methods are further proposed and evaluated on the same dataset under the same task settings following the task design of \cite{yu2019uncertainty}. Specifically, several researches further promote the benchmark with 10\% (i.e. 8 scans) labeled scans to further evaluate the performance under the circumstance with fewer labeled data.

\textbf{Pancreas CT dataset. }
The NIH Pancreas CT segmentation dataset \cite{clark2013cancer} contains 82 3D abdominal contrast-enhanced CT volumes, which are collected from
53 male and 27 female subjects at the National Institutes of Health Clinical Center \footnote{https://wiki.cancerimagingarchive.net/display/Public/Pancreas-CT}.
The dataset are collected on Philips and Siemens MDCT scanners and have a fixed resolution of $512 \times 512$ with varying thicknesses from 1.5 to 2.5 mm, while the axial view slice number can vary from 181 to 466. In \cite{xia20203d}, the dataset is randomly split into 20 testing cases and 62 training cases. Experimental results with 10\% labeled training cases (6 labeled and 56 unlabeled) and 20\% labeled training cases (12 labeled and 50 unlabeled) is reported.
Following the pre-processing in \cite{zhou2019prior}, the voxel values are clipped to the range of [-125,275] Hounsfield Units (HU) and further re-sampled to an isotropic resolution of $1 \times 1 \times 1 mm^{3}$.
several researches further promote the benchmark with 10\% (i.e. 8 scans) labeled scans to further evaluate the performance under the circumstance with fewer labeled data.
Several semi-supervised approaches \cite{luo2022semi,wu2022mutual,9740182} are evaluated on Pancreas CT dataset.

\textbf{BraTS dataset. }
The Brain Tumor Segmentation (BraTS) 2019 dataset \cite{menze2014multimodal} contains multi-institutional preoperative MRI of 335 glioma patients, where each patient has four modalities of MRI scans including T1, T1Gd, T2 and T2-FLAIR with neuroradiologist-examined labels. For several existing approaches \cite{9741294,zhang2023uncertainty,luo2022semi}, T2-FLAIR for whole tumor segmentation is used since such modality can better manifest the malignant tumors \cite{zeineldin2020deepseg}. All the scans are re-sampled to the same resolution of $1 \times 1 \times 1 mm^{3}$ with intensity normalized to zero mean and unit variance.
For semi-supervised settings, the dataset is splitted into 250 scans for training, 25 scans for validation and the remaining 60
scans for testing. Among the 250 training scans, two different settings are performed with 10\%/25 and 20\%/50 scans as labeled data and the remaining scans as unlabeled data.

\textbf{ACDC dataset. }
The ACDC (Automated Cardiac Diagnosis Challenge) dataset \cite{bernard2018deep} was collected from real clinical exams acquired at the University Hospital of Dijon \footnote{https://www.creatis.insa-lyon.fr/Challenge/acdc/databases.html}.
The dataset contains multi-slice 2D cine cardiac MR imaging samples from 100 patients for training. For semi-supervised settings, the dataset is splitted into 70 scans for training, 10 scans for validation and 20 scans for testing. Unlike previous 3D binary segmentation benchmark datasets, ACDC is a 2D multi-class segmentation task including RV cavity, myocardium and the LV cavity.

\section{Existing Challenges and Future Directions }

Although considerable performance has been achieved for semi-supervised medical image segmentation tasks, there are still several open questions for future work. In this section, we outline some of these challenges and potential future directions as follows.

\textbf{Misaligned distribution and class imbalance.}
As described in Section \ref{sec:benchmark}, existing semi-supervised medical image segmentation approaches have achieved comparable results with upper-bound fully supervised results in some benchmark datasets like LA segmentation \cite{xiong2021global}. However, these benchmarks are relatively "simple" tasks, with small amount of experimental data where the training and test set are from the same domain/medical center. 
However, a clinical applicable deep learning model should be generalized suitably across multiple centres and scanner vendors from different domains \cite{campello2021multi}.
As there is usually a large amount of unlabeled data in semi-supervised learning, the distribution of labeled and unlabeled data may be misaligned. This limitation is also highlighted by recent semi-supervised medical segmentation benchmarks like \cite{ma2021abdomenct} and FLARE 22 challenge \footnote{https://flare22.grand-challenge.org}. Based on the work in \cite{oliver2018realistic}, adding unlabeled data from a mismatched distribution from labeled data can lower the performance compared to not using any unlabeled data. Therefore, it is of great importance to issue the challenge of misaligned distribution for semi-supervised learning. 
As for class imbalance, when the training data is highly imbalanced, the trained model will show bias towards the majority classes, and may completely ignore the minority classes in some extreme cases \cite{johnson2019survey}.
Besides, for semi-supervised multi-class segmentation, there usually exists the uncertainty imbalance problem brought by class imbalance and limited labeled data. Recent studies \cite{van2022pitfalls} found that aleatoric uncertainty derived from the entropy of the predictions may lead to sub-optimal results in a multi-class context.

\textbf{Methodological analysis.} Existing semi-supervised medical image segmentation approaches predominantly use unlabeled data to generate constraints, then the models are updated with supervised loss for labeled data and unsupervised loss/constraints for unlabeled data (or both labeled and unlabeled data). 
Generally, there is only a single weight to balance between supervised and unsupervised loss as described in many approaches \cite{yu2019uncertainty,luo2021semi,zhang2021dual}. In other words, all the unlabeled data are treated equally for semi-supervised learning.
However, not all unlabeled data is equally appropriate for the learning procedure of the model.
For example, when the estimation of an unlabeled case is incorrect, training on that particular label-estimate may hurt the overall performance. 
To issue this problem, it is important to encourage the model focusing on more challenging areas/cases and therefore exploit more useful information from unlabeled data like assigning different weights for each unlabeled example \cite{Ren2020NotAU}. 
Recent studies \cite{ghosh2020data} also found that the quality of the perturbations is key to obtaining reasonable performances for semi-supervised learning, especially in the case of efficient data augmentations or perturbations schemes when the data lies in the neighborhood of low-dimensional manifolds.

\textbf{Integration with other annotation-efficient approaches.}
For existing semi-supervised learning approaches, we still need a small amount of well-annotated labeled data to guide the learning of unlabeled data.
However, acquiring such fully annotated training data can still be costly, especially for the tasks of medical image segmentation.
To further alleviate the annotation cost, some researches integrate semi-supervised learning with other annotation-efficient approaches like utilizing partially labelled datasets \cite{zhang2021automatic}, leveraging image-level, box-level and pixel-level annotations \cite{reiss2021every} or scribble supervisions \cite{zhang2022cyclemix}, or exploiting noisy labeled data \cite{xu2021noisy}. 
Semi-supervised medical image segmentation could also be integrated with few-shot segmentation to improve the generalization ability with combination strategies to segment similar objects in unseen images. Both methods aim to improve the performance of a model when there is limited labeled data available. In semi-supervised learning, the model learns from both the labeled and unlabeled data by making assumptions about the distribution of the data, which is different from few-shot learning. Besides, with the recent introduction of SAM \cite{SAM-Meta}, which can serve as pseudo-label generator for image segmentation \cite{jiang2023segment,li2023segment}, may provide some insights into the future development of semi-supervised learning for medical image segmentation \cite{SAM4MIS}.

\section{Conclusion}

Semi-supervised learning has been widely applied to medical image segmentation tasks since it alleviates the
heavy burden of acquiring expert-examined annotations and takes the advantage of unlabeled data which is
much easier to acquire.
In this survey, we provide a taxonomy of existing deep semi-supervised learning methods for medical image segmentation tasks and group these methods into three main categories, namely, pseudo labels, unsupervised regularization, and knowledge priors.  
Other than summarizing technical novelties of these approaches, we also analyse and discuss the empirical results of these methods on several public benchmark datasets. Furthermore, we analysed and discussed the limitations and several unsolved problems of existing approaches.
We hope this review could inspire the research community to explore solutions for this challenge and further promote the developments in this impactful area of research.


\section*{Acknowledgment}
This paper was supported in part by the National Science Foundation under Grant 32000687, and in part by the University Synergy Innovation Program of Anhui Province
under Grant GXXT-2019-044 and Beijing Natural Science Foundation under Grant Z200024. We also appreciate the efforts of literature collection and code implementations of SSL4MIS \footnote{https://github.com/HiLab-git/SSL4MIS} and several public benchmarks.

\bibliographystyle{model2-names}
\bibliography{reference}

\end{document}